\pdfoutput=1
\documentclass[conference]{IEEEtran}
\IEEEoverridecommandlockouts

\usepackage[T1]{fontenc}
\usepackage[utf8]{inputenc}
\usepackage{amsmath}
\usepackage{newtxtext}
\usepackage{newtxmath}
\usepackage{graphicx}
\usepackage{booktabs}
\usepackage{tikz}
\usepackage{array}
\usepackage{xcolor}
\usepackage{microtype}
\usepackage[numbers,sort&compress]{natbib}
\usepackage{url}
\usepackage[hidelinks]{hyperref}
\hypersetup{
  pdftitle={Component Translation and Judge-Labelled Refusal: A Multilingual Safety Evaluation},
  pdfauthor={Yongxi Zhou, Junwei Yao, Wenbo Ye, Jiahui Ge},
  pdfsubject={Multilingual safety evaluation through complete-component translation},
  pdfkeywords={large language models, safety evaluation, multilingual, code-switching,
    measurement validity, factorial design, LLM-as-judge}}

\usepackage{CJKutf8}
\newcommand{\zh}[1]{\begin{CJK*}{UTF8}{gbsn}#1\end{CJK*}}

\newcommand{\cfg}[1]{\texttt{#1}}
\newcommand{\pp}{\,pp}

\makeatletter
\renewcommand{\paragraph}[1]{\par\vspace{2pt}\noindent\textbf{#1}\ \ignorespaces}
\makeatother

\newcommand{\RefusalFamily}{46}

\newcommand{\RefusalDsRequestZhRd}{\ensuremath{-0.011}}

\newcommand{\RefusalDsOperationEsRd}{\ensuremath{-0.063}}

\newcommand{\RefusalDsOperationZhRd}{\ensuremath{-0.014}}

\newcommand{\RefusalGemRequestEsRd}{\ensuremath{-0.023}}

\newcommand{\RefusalGemOperationEsRd}{\ensuremath{-0.067}}

\newcommand{\RefusalGemOperationZhRd}{\ensuremath{-0.083}}

\newcommand{\RefusalDsIxEsxesEst}{\ensuremath{+0.047}}
\newcommand{\RefusalDsIxEsxesCi}{\ensuremath{[+0.021, +0.079]}}
\newcommand{\RefusalDsIxEsxesPholm}{\ensuremath{0.016}}

\newcommand{\RefusalDsSwapSRd}{\ensuremath{+5.3}}
\newcommand{\RefusalDsSwapSCi}{\ensuremath{[+1.6, +8.9]}}
\newcommand{\RefusalDsSwapSP}{\ensuremath{0.013}}
\newcommand{\RefusalDsSwapSPpre}{\ensuremath{0.065}}
\newcommand{\RefusalDsSwapSDisc}{12:2}
\newcommand{\RefusalDsSwapZRd}{\ensuremath{-0.5}}
\newcommand{\RefusalDsSwapZCi}{\ensuremath{[-2.6, +1.6]}}

\newcommand{\RefusalDsSwapZPpre}{\ensuremath{1.000}}
\newcommand{\RefusalDsSwapZDisc}{2:3}
\newcommand{\RefusalGemSwapSRd}{\ensuremath{+5.3}}
\newcommand{\RefusalGemSwapSCi}{\ensuremath{[+1.6, +9.1]}}
\newcommand{\RefusalGemSwapSP}{\ensuremath{0.013}}
\newcommand{\RefusalGemSwapSPpre}{\ensuremath{0.065}}
\newcommand{\RefusalGemSwapSDisc}{12:2}
\newcommand{\RefusalGemSwapZRd}{\ensuremath{+5.3}}
\newcommand{\RefusalGemSwapZCi}{\ensuremath{[+1.1, +9.5]}}
\newcommand{\RefusalGemSwapZP}{\ensuremath{0.021}}
\newcommand{\RefusalGemSwapZPpre}{\ensuremath{0.065}}
\newcommand{\RefusalGemSwapZDisc}{13:3}
\newcommand{\RefusalDsTransferRd}{\ensuremath{+5.8}}
\newcommand{\RefusalDsTransferCi}{\ensuremath{[+2.1, +9.5]}}
\newcommand{\RefusalDsTransferPpre}{\ensuremath{0.002}}
\newcommand{\RefusalGemTransferRd}{\ensuremath{+0.0}}
\newcommand{\RefusalGemTransferCi}{\ensuremath{[-4.9, +5.4]}}
\newcommand{\RefusalGemTransferPpre}{\ensuremath{1.000}}

\newcommand{\RefusalgptminiGptminSwapSRd}{\ensuremath{+0.5}}
\newcommand{\RefusalgptminiGptminSwapSCi}{\ensuremath{[+0.0, +1.6]}}

\newcommand{\RefusalgptminiGptminSwapSPpre}{\ensuremath{1.000}}
\newcommand{\RefusalgptminiGptminSwapSDisc}{1:0}
\newcommand{\RefusalgptminiGptminSwapZRd}{\ensuremath{+1.6}}
\newcommand{\RefusalgptminiGptminSwapZCi}{\ensuremath{[-0.5, +4.2]}}

\newcommand{\RefusalgptminiGptminSwapZPpre}{\ensuremath{1.000}}
\newcommand{\RefusalgptminiGptminSwapZDisc}{4:1}
\newcommand{\RefusalgptminiGptminTransferRd}{\ensuremath{-1.1}}
\newcommand{\RefusalgptminiGptminTransferCi}{\ensuremath{[-3.7, +1.6]}}

\newcommand{\RefusalgptminiqwenGptminSwapSRd}{\ensuremath{+0.0}}
\newcommand{\RefusalgptminiqwenGptminSwapSCi}{\ensuremath{[+0.0, +0.0]}}

\newcommand{\RefusalgptminiqwenGptminSwapSPpre}{\ensuremath{1.000}}

\newcommand{\RefusalgptminiqwenGptminSwapZRd}{\ensuremath{+0.0}}
\newcommand{\RefusalgptminiqwenGptminSwapZCi}{\ensuremath{[-2.1, +2.1]}}

\newcommand{\RefusalgptminiqwenGptminSwapZPpre}{\ensuremath{1.000}}

\newcommand{\RefusalgptminiqwenGptminTransferRd}{\ensuremath{+0.0}}
\newcommand{\RefusalgptminiqwenGptminTransferCi}{\ensuremath{[-2.1, +2.1]}}

\newcommand{\RefusalmistralsmallMistraSwapSRd}{\ensuremath{+13.7}}
\newcommand{\RefusalmistralsmallMistraSwapSCi}{\ensuremath{[+8.4, +18.9]}}

\newcommand{\RefusalmistralsmallMistraSwapSPpre}{\ensuremath{<0.001}}
\newcommand{\RefusalmistralsmallMistraSwapSDisc}{28:2}
\newcommand{\RefusalmistralsmallMistraSwapZRd}{\ensuremath{+2.6}}
\newcommand{\RefusalmistralsmallMistraSwapZCi}{\ensuremath{[-3.2, +8.4]}}

\newcommand{\RefusalmistralsmallMistraSwapZPpre}{\ensuremath{0.473}}
\newcommand{\RefusalmistralsmallMistraSwapZDisc}{18:13}
\newcommand{\RefusalmistralsmallMistraTransferRd}{\ensuremath{+11.1}}
\newcommand{\RefusalmistralsmallMistraTransferCi}{\ensuremath{[+4.7, +17.9]}}

\newcommand{\RefusalmistralsmallqwenMistraSwapSRd}{\ensuremath{+12.1}}
\newcommand{\RefusalmistralsmallqwenMistraSwapSCi}{\ensuremath{[+6.8, +17.4]}}

\newcommand{\RefusalmistralsmallqwenMistraSwapSPpre}{\ensuremath{<0.001}}

\newcommand{\RefusalmistralsmallqwenMistraSwapZRd}{\ensuremath{+2.1}}
\newcommand{\RefusalmistralsmallqwenMistraSwapZCi}{\ensuremath{[-3.7, +7.4]}}

\newcommand{\RefusalmistralsmallqwenMistraSwapZPpre}{\ensuremath{0.572}}

\newcommand{\RefusalmistralsmallqwenMistraTransferRd}{\ensuremath{+10.0}}
\newcommand{\RefusalmistralsmallqwenMistraTransferCi}{\ensuremath{[+4.2, +16.3]}}

\newcommand{\RefusalqwenDsSwapSRd}{\ensuremath{+5.3}}
\newcommand{\RefusalqwenDsSwapSCi}{\ensuremath{[+2.1, +8.9]}}

\newcommand{\RefusalqwenDsSwapSPpre}{\ensuremath{0.025}}

\newcommand{\RefusalqwenDsSwapZRd}{\ensuremath{-0.5}}
\newcommand{\RefusalqwenDsSwapZCi}{\ensuremath{[-2.6, +1.6]}}

\newcommand{\RefusalqwenDsSwapZPpre}{\ensuremath{1.000}}

\newcommand{\RefusalqwenGemSwapSRd}{\ensuremath{+5.9}}
\newcommand{\RefusalqwenGemSwapSCi}{\ensuremath{[+2.6, +9.6]}}

\newcommand{\RefusalqwenGemSwapSPpre}{\ensuremath{0.017}}

\newcommand{\RefusalqwenGemSwapZRd}{\ensuremath{+4.8}}
\newcommand{\RefusalqwenGemSwapZCi}{\ensuremath{[+1.1, +8.7]}}

\newcommand{\RefusalqwenGemSwapZPpre}{\ensuremath{0.105}}

\newcommand{\RefusalqwenDsTransferRd}{\ensuremath{+5.8}}
\newcommand{\RefusalqwenDsTransferCi}{\ensuremath{[+2.6, +9.5]}}

\newcommand{\RefusalqwenGemTransferRd}{\ensuremath{+1.1}}
\newcommand{\RefusalqwenGemTransferCi}{\ensuremath{[-3.8, +5.9]}}

\newcommand{\UnsafeDsRequestEsRd}{\ensuremath{+0.016}}

\newcommand{\UnsafeDsRequestZhRd}{\ensuremath{+0.009}}

\newcommand{\UnsafeDsOperationEsRd}{\ensuremath{+0.056}}

\newcommand{\UnsafeGemOperationZhRd}{\ensuremath{+0.065}}

\newcommand{\UnsafeDsSwapSRd}{\ensuremath{-4.7}}
\newcommand{\UnsafeDsSwapSCi}{\ensuremath{[-8.4, -1.6]}}

\newcommand{\UnsafeDsSwapSPpre}{\ensuremath{0.047}}

\newcommand{\UnsafeDsSwapZRd}{\ensuremath{+0.0}}
\newcommand{\UnsafeDsSwapZCi}{\ensuremath{[-2.6, +2.6]}}

\newcommand{\UnsafeDsSwapZPpre}{\ensuremath{1.000}}

\newcommand{\UnsafeGemSwapSRd}{\ensuremath{-5.9}}
\newcommand{\UnsafeGemSwapSCi}{\ensuremath{[-9.7, -2.1]}}

\newcommand{\UnsafeGemSwapSPpre}{\ensuremath{0.037}}

\newcommand{\UnsafeGemSwapZRd}{\ensuremath{-3.7}}
\newcommand{\UnsafeGemSwapZCi}{\ensuremath{[-7.4, -0.5]}}

\newcommand{\UnsafeGemSwapZPpre}{\ensuremath{0.196}}

\newcommand{\UnsafeDsTransferRd}{\ensuremath{-4.7}}
\newcommand{\UnsafeDsTransferCi}{\ensuremath{[-8.4, -1.6]}}

\newcommand{\UnsafeGemTransferRd}{\ensuremath{-2.2}}
\newcommand{\UnsafeGemTransferCi}{\ensuremath{[-7.6, +2.7]}}

\newcommand{\UnsafegptminiGptminSwapSRd}{\ensuremath{-0.5}}
\newcommand{\UnsafegptminiGptminSwapSCi}{\ensuremath{[-1.6, +0.0]}}

\newcommand{\UnsafegptminiGptminSwapSPpre}{\ensuremath{1.000}}

\newcommand{\UnsafegptminiGptminSwapZRd}{\ensuremath{-1.1}}
\newcommand{\UnsafegptminiGptminSwapZCi}{\ensuremath{[-2.6, +0.0]}}

\newcommand{\UnsafegptminiGptminSwapZPpre}{\ensuremath{1.000}}

\newcommand{\UnsafegptminiGptminTransferRd}{\ensuremath{+0.5}}
\newcommand{\UnsafegptminiGptminTransferCi}{\ensuremath{[-1.1, +2.6]}}

\newcommand{\UnsafegptminiqwenGptminSwapSRd}{\ensuremath{+0.0}}
\newcommand{\UnsafegptminiqwenGptminSwapSCi}{\ensuremath{[+0.0, +0.0]}}

\newcommand{\UnsafegptminiqwenGptminSwapSPpre}{\ensuremath{1.000}}

\newcommand{\UnsafegptminiqwenGptminSwapZRd}{\ensuremath{-1.1}}
\newcommand{\UnsafegptminiqwenGptminSwapZCi}{\ensuremath{[-2.6, +0.0]}}

\newcommand{\UnsafegptminiqwenGptminSwapZPpre}{\ensuremath{1.000}}

\newcommand{\UnsafegptminiqwenGptminTransferRd}{\ensuremath{+1.1}}
\newcommand{\UnsafegptminiqwenGptminTransferCi}{\ensuremath{[+0.0, +2.6]}}

\newcommand{\UnsafemistralsmallMistraSwapSRd}{\ensuremath{-11.1}}
\newcommand{\UnsafemistralsmallMistraSwapSCi}{\ensuremath{[-16.3, -6.3]}}

\newcommand{\UnsafemistralsmallMistraSwapSPpre}{\ensuremath{<0.001}}

\newcommand{\UnsafemistralsmallMistraSwapZRd}{\ensuremath{-0.5}}
\newcommand{\UnsafemistralsmallMistraSwapZCi}{\ensuremath{[-6.3, +5.3]}}

\newcommand{\UnsafemistralsmallMistraSwapZPpre}{\ensuremath{1.000}}

\newcommand{\UnsafemistralsmallMistraTransferRd}{\ensuremath{-10.5}}
\newcommand{\UnsafemistralsmallMistraTransferCi}{\ensuremath{[-17.4, -4.2]}}

\newcommand{\UnsafemistralsmallqwenMistraSwapSRd}{\ensuremath{-12.1}}
\newcommand{\UnsafemistralsmallqwenMistraSwapSCi}{\ensuremath{[-17.4, -6.8]}}

\newcommand{\UnsafemistralsmallqwenMistraSwapSPpre}{\ensuremath{<0.001}}

\newcommand{\UnsafemistralsmallqwenMistraSwapZRd}{\ensuremath{-1.6}}
\newcommand{\UnsafemistralsmallqwenMistraSwapZCi}{\ensuremath{[-7.4, +4.2]}}

\newcommand{\UnsafemistralsmallqwenMistraSwapZPpre}{\ensuremath{0.711}}

\newcommand{\UnsafemistralsmallqwenMistraTransferRd}{\ensuremath{-10.5}}
\newcommand{\UnsafemistralsmallqwenMistraTransferCi}{\ensuremath{[-16.8, -4.2]}}

\newcommand{\UnsafeqwenDsSwapSRd}{\ensuremath{-5.3}}
\newcommand{\UnsafeqwenDsSwapSCi}{\ensuremath{[-8.9, -2.1]}}

\newcommand{\UnsafeqwenDsSwapSPpre}{\ensuremath{0.025}}

\newcommand{\UnsafeqwenDsSwapZRd}{\ensuremath{+0.5}}
\newcommand{\UnsafeqwenDsSwapZCi}{\ensuremath{[-1.6, +2.6]}}

\newcommand{\UnsafeqwenDsSwapZPpre}{\ensuremath{1.000}}

\newcommand{\UnsafeqwenGemSwapSRd}{\ensuremath{-6.9}}
\newcommand{\UnsafeqwenGemSwapSCi}{\ensuremath{[-10.7, -3.2]}}

\newcommand{\UnsafeqwenGemSwapSPpre}{\ensuremath{0.005}}

\newcommand{\UnsafeqwenGemSwapZRd}{\ensuremath{-4.8}}
\newcommand{\UnsafeqwenGemSwapZCi}{\ensuremath{[-8.6, -1.1]}}

\newcommand{\UnsafeqwenGemSwapZPpre}{\ensuremath{0.067}}

\newcommand{\UnsafeqwenDsTransferRd}{\ensuremath{-5.8}}
\newcommand{\UnsafeqwenDsTransferCi}{\ensuremath{[-9.5, -2.6]}}

\newcommand{\UnsafeqwenGemTransferRd}{\ensuremath{-2.2}}
\newcommand{\UnsafeqwenGemTransferCi}{\ensuremath{[-7.6, +2.7]}}

{\textbf{[results.tex not built]}}%
  }

\newcommand{\AppCoverageTable}{%
\begin{tabular}{@{}lrrrr@{}}
\toprule
source-native group & pool & two slots & qualifier & frame content word \\
\midrule
HarmBench: copyright & 53 & 53 & 39 & 0 \\
chemical / biological & 28 & 27 & 22 & 21 \\
harmful & 21 & 17 & 6 & 15 \\
illegal & 59 & 58 & 43 & 49 \\
misinformation / disinformation & 41 & 41 & 27 & 33 \\
harassment / bullying & 25 & 24 & 8 & 18 \\
cybercrime / intrusion & 61 & 60 & 28 & 49 \\
AdvBench & 82 & 81 & 17 & 73 \\
\midrule
total & 370 & 361 & 190 & 258 \\
\bottomrule
\end{tabular}}

\newcommand{\AppCrosswalkTable}{%
\begin{tabular}{@{}llrrr@{}}
\toprule
inherited stratum & source-native group & sampled & consensus & analysed \\
\midrule
\texttt{cyber\_abuse} & cybercrime / intrusion & 48 & 40 & 39 \\
\texttt{cyber\_abuse} & AdvBench & 33 & 32 & 29 \\
\texttt{fraud\_deception} & illegal & 48 & 42 & 40 \\
\texttt{fraud\_deception} & misinformation / disinformation & 34 & 25 & 20 \\
\texttt{physical\_harm} & HarmBench: copyright & 44 & 27 & 0 \\
\texttt{physical\_harm} & chemical / biological & 20 & 17 & 3 \\
\texttt{physical\_harm} & harmful & 15 & 13 & 12 \\
\texttt{violent\_wrongdoing} & HarmBench: copyright & 1 & 1 & 0 \\
\texttt{violent\_wrongdoing} & harassment / bullying & 21 & 20 & 20 \\
\texttt{violent\_wrongdoing} & AdvBench & 36 & 28 & 27 \\
\midrule
total & & 300 & 245 & 190 \\
\bottomrule
\end{tabular}}

\newcommand{\AppGateTable}{%
\begin{tabular}{@{}lrrrr@{}}
\toprule
configuration & intent: negative & intent: missing & naturalness 1 & naturalness unrated \\
\midrule
\texttt{EE} & 24 & 15 & 5 & 15 \\
\texttt{EZ} & 20 & 13 & 0 & 3 \\
\texttt{ES} & 27 & 11 & 1 & 3 \\
\texttt{ZE} & 25 & 12 & 0 & 2 \\
\texttt{ZS} & 30 & 14 & 0 & 4 \\
\texttt{ZZ} & 17 & 16 & 12 & 16 \\
\texttt{SE} & 23 & 11 & 0 & 7 \\
\texttt{SZ} & 22 & 14 & 1 & 13 \\
\texttt{SS} & 31 & 11 & 2 & 9 \\
\midrule
total & 219 & 117 & 21 & 72 \\
\bottomrule
\end{tabular}}
\newcommand{\AppGateHarmfulSeeds}{245}

\newcommand{\AppGateQMissing}{21.4}
\newcommand{\AppGatePMissing}{0.006}
\newcommand{\AppGateQNegative}{34.9}
\newcommand{\AppGatePNegative}{\ensuremath{2.7\times 10^{-5}}}

\newcommand{\AppGateCanonicalNegatives}{24}

\newcommand{\AppRoundtripTable}{%
\begin{tabular}{@{}lrrrrrrr@{}}
\toprule
run & pairs & sentence & spans & difference & 95\% interval & Wilcoxon $p$ & spans worse \\
\midrule
replay, zh & 305 & 0.794 & 0.787 & \ensuremath{-0.007} & \ensuremath{[-0.024, +0.011]} & 0.93 & 137 \\
replay, es & 296 & 0.895 & 0.887 & \ensuremath{-0.008} & \ensuremath{[-0.020, +0.003]} & 0.25 & 142 \\
original, zh & 305 & 0.794 & 0.787 & \ensuremath{-0.007} & \ensuremath{[-0.025, +0.009]} & 0.93 & 137 \\
original, es & 305 & 0.894 & 0.887 & \ensuremath{-0.007} & \ensuremath{[-0.019, +0.005]} & 0.30 & 145 \\
\bottomrule
\end{tabular}}
\newcommand{\AppRoundtripResamples}{5{,}000}
\newcommand{\AppRoundtripSeed}{20261025}
\newcommand{\AppRoundtripMissingEs}{9}

\newcommand{\AppAuditTable}{%
\begin{tabular}{@{}lrrrrrlr@{}}
\toprule
comparison & YY & YN & NY & NN & agree (\%) & $\kappa$ [cluster 95\%] & weighted $\kappa$ \\
\midrule
rater A vs rater B, refusal & 163 & 1 & 1 & 20 & 98.9 & 0.946 [0.86, 1.00] & 0.946 \\
rater A vs rater B, unsafe compliance & 14 & 0 & 1 & 168 & 99.5 & 0.963 [0.85, 1.00] & 0.963 \\
\midrule
reference vs primary judge, refusal & 162 & 2 & 1 & 20 & 98.4 & 0.921 [0.82, 1.00] & 0.920 \\
reference vs second judge, refusal & 163 & 1 & 1 & 20 & 98.9 & 0.946 [0.86, 1.00] & 0.946 \\
reference vs primary judge, unsafe compliance & 14 & 1 & 6 & 164 & 96.2 & 0.780 [0.59, 0.91] & 0.783 \\
reference vs second judge, unsafe compliance & 14 & 1 & 7 & 163 & 95.7 & 0.755 [0.57, 0.89] & 0.758 \\
\bottomrule
\end{tabular}}
\newcommand{\AppAuditItems}{185}
\newcommand{\AppAuditClusters}{118}
\newcommand{\AppAuditStrata}{36}
\newcommand{\AppAuditPopulation}{6{,}830}
\newcommand{\AppAuditResamples}{10{,}000}
\newcommand{\AppAuditSeed}{20261026}
\newcommand{\AppAuditAdjudicated}{5}
\newcommand{\AppAuditAdjudicatedItems}{4}
\newcommand{\AppAuditRefUnsafeYes}{15}
\newcommand{\AppAuditRefRefusalYes}{164}

\newcommand{\AppUnsafeCells}{%
\begin{tabular}{@{}lrrrr@{}}
\toprule
config. & \multicolumn{1}{c}{Gemini 2.5 Flash} & \multicolumn{1}{c}{DeepSeek} & \multicolumn{1}{c}{GPT-5.4-mini} & \multicolumn{1}{c}{Mistral Small 4} \\
\midrule
\texttt{EE} & \ensuremath{3.2} [\ensuremath{1.5}, \ensuremath{6.7}] & \ensuremath{2.1} [\ensuremath{0.8}, \ensuremath{5.3}] & \ensuremath{0.0} [\ensuremath{0.0}, \ensuremath{2.0}] & \ensuremath{11.1} [\ensuremath{7.3}, \ensuremath{16.3}] \\
\addlinespace[2pt]
\texttt{ES} & \ensuremath{9.6} [\ensuremath{6.1}, \ensuremath{14.6}] & \ensuremath{10.0} [\ensuremath{6.5}, \ensuremath{15.1}] & \ensuremath{0.5} [\ensuremath{0.1}, \ensuremath{2.9}] & \ensuremath{22.6} [\ensuremath{17.3}, \ensuremath{29.1}] \\
\texttt{EZ} & \ensuremath{9.1} [\ensuremath{5.8}, \ensuremath{14.1}] & \ensuremath{3.7} [\ensuremath{1.8}, \ensuremath{7.4}] & \ensuremath{1.1} [\ensuremath{0.3}, \ensuremath{3.8}] & \ensuremath{19.5} [\ensuremath{14.5}, \ensuremath{25.7}] \\
\texttt{SE} & \ensuremath{3.7} [\ensuremath{1.8}, \ensuremath{7.4}] & \ensuremath{5.3} [\ensuremath{2.9}, \ensuremath{9.4}] & \ensuremath{0.0} [\ensuremath{0.0}, \ensuremath{2.0}] & \ensuremath{11.6} [\ensuremath{7.8}, \ensuremath{16.9}] \\
\texttt{ZE} & \ensuremath{5.3} [\ensuremath{2.9}, \ensuremath{9.4}] & \ensuremath{3.7} [\ensuremath{1.8}, \ensuremath{7.4}] & \ensuremath{0.0} [\ensuremath{0.0}, \ensuremath{2.0}] & \ensuremath{18.9} [\ensuremath{14.0}, \ensuremath{25.1}] \\
\addlinespace[2pt]
\texttt{SS} & \ensuremath{10.1} [\ensuremath{6.5}, \ensuremath{15.2}] & \ensuremath{8.9} [\ensuremath{5.7}, \ensuremath{13.9}] & \ensuremath{0.0} [\ensuremath{0.0}, \ensuremath{2.0}] & \ensuremath{21.6} [\ensuremath{16.3}, \ensuremath{28.0}] \\
\texttt{SZ} & \ensuremath{11.6} [\ensuremath{7.8}, \ensuremath{17.0}] & \ensuremath{6.3} [\ensuremath{3.6}, \ensuremath{10.7}] & \ensuremath{0.0} [\ensuremath{0.0}, \ensuremath{2.0}] & \ensuremath{18.4} [\ensuremath{13.6}, \ensuremath{24.5}] \\
\texttt{ZS} & \ensuremath{9.6} [\ensuremath{6.1}, \ensuremath{14.6}] & \ensuremath{8.9} [\ensuremath{5.7}, \ensuremath{13.9}] & \ensuremath{0.0} [\ensuremath{0.0}, \ensuremath{2.0}] & \ensuremath{25.3} [\ensuremath{19.6}, \ensuremath{31.9}] \\
\texttt{ZZ} & \ensuremath{11.1} [\ensuremath{7.4}, \ensuremath{16.4}] & \ensuremath{5.8} [\ensuremath{3.3}, \ensuremath{10.1}] & \ensuremath{0.0} [\ensuremath{0.0}, \ensuremath{2.0}] & \ensuremath{23.7} [\ensuremath{18.2}, \ensuremath{30.2}] \\
\bottomrule
\end{tabular}}

\newcommand{\AppRefusalSecondCells}{%
\begin{tabular}{@{}lrrrr@{}}
\toprule
config. & \multicolumn{1}{c}{Gemini 2.5 Flash} & \multicolumn{1}{c}{DeepSeek} & \multicolumn{1}{c}{GPT-5.4-mini} & \multicolumn{1}{c}{Mistral Small 4} \\
\midrule
\texttt{EE} & \ensuremath{96.3} [\ensuremath{92.6}, \ensuremath{98.2}] & \ensuremath{97.4} [\ensuremath{94.0}, \ensuremath{98.9}] & \ensuremath{99.5} [\ensuremath{97.1}, \ensuremath{99.9}] & \ensuremath{88.9} [\ensuremath{83.7}, \ensuremath{92.7}] \\
\addlinespace[2pt]
\texttt{ES} & \ensuremath{89.4} [\ensuremath{84.1}, \ensuremath{93.0}] & \ensuremath{89.5} [\ensuremath{84.3}, \ensuremath{93.1}] & \ensuremath{99.5} [\ensuremath{97.1}, \ensuremath{99.9}] & \ensuremath{76.3} [\ensuremath{69.8}, \ensuremath{81.8}] \\
\texttt{EZ} & \ensuremath{88.2} [\ensuremath{82.8}, \ensuremath{92.1}] & \ensuremath{96.3} [\ensuremath{92.6}, \ensuremath{98.2}] & \ensuremath{98.9} [\ensuremath{96.2}, \ensuremath{99.7}] & \ensuremath{79.5} [\ensuremath{73.2}, \ensuremath{84.6}] \\
\texttt{SE} & \ensuremath{95.3} [\ensuremath{91.2}, \ensuremath{97.5}] & \ensuremath{94.7} [\ensuremath{90.6}, \ensuremath{97.1}] & \ensuremath{99.5} [\ensuremath{97.1}, \ensuremath{99.9}] & \ensuremath{88.4} [\ensuremath{83.1}, \ensuremath{92.2}] \\
\texttt{ZE} & \ensuremath{93.2} [\ensuremath{88.6}, \ensuremath{96.0}] & \ensuremath{95.8} [\ensuremath{91.9}, \ensuremath{97.9}] & \ensuremath{98.9} [\ensuremath{96.2}, \ensuremath{99.7}] & \ensuremath{81.6} [\ensuremath{75.5}, \ensuremath{86.4}] \\
\addlinespace[2pt]
\texttt{SS} & \ensuremath{88.4} [\ensuremath{83.0}, \ensuremath{92.2}] & \ensuremath{90.0} [\ensuremath{84.9}, \ensuremath{93.5}] & \ensuremath{99.5} [\ensuremath{97.1}, \ensuremath{99.9}] & \ensuremath{77.9} [\ensuremath{71.5}, \ensuremath{83.2}] \\
\texttt{SZ} & \ensuremath{85.7} [\ensuremath{80.0}, \ensuremath{90.0}] & \ensuremath{93.7} [\ensuremath{89.3}, \ensuremath{96.4}] & \ensuremath{98.9} [\ensuremath{96.2}, \ensuremath{99.7}] & \ensuremath{79.5} [\ensuremath{73.2}, \ensuremath{84.6}] \\
\texttt{ZS} & \ensuremath{88.8} [\ensuremath{83.5}, \ensuremath{92.6}] & \ensuremath{90.0} [\ensuremath{84.9}, \ensuremath{93.5}] & \ensuremath{99.5} [\ensuremath{97.1}, \ensuremath{99.9}] & \ensuremath{74.7} [\ensuremath{68.1}, \ensuremath{80.4}] \\
\texttt{ZZ} & \ensuremath{87.3} [\ensuremath{81.8}, \ensuremath{91.3}] & \ensuremath{94.2} [\ensuremath{89.9}, \ensuremath{96.7}] & \ensuremath{98.4} [\ensuremath{95.5}, \ensuremath{99.5}] & \ensuremath{75.8} [\ensuremath{69.2}, \ensuremath{81.3}] \\
\bottomrule
\end{tabular}}

}

\begin{document}

\title{Component Translation and Judge-Labelled Refusal: A Multilingual Safety Evaluation}

\author{%
\IEEEauthorblockN{Yongxi Zhou, Junwei Yao, Wenbo Ye, Jiahui Ge%
\thanks{Preprint. Appendices~\ref{app:coverage} to~\ref{app:cells} extend the ten-page
manuscript prepared for the 4th Workshop on Large Language and Foundation Models (WLLFM 2026)
at IEEE BigData 2026; the main text is otherwise that manuscript. A replication archive
accompanies this preprint as ancillary files.}}
\IEEEauthorblockA{Independent Researcher\\
Corresponding author: Yongxi Zhou (zhou.yongx@northeastern.edu)}}

\maketitle

\begin{abstract}
Whole-prompt translation evaluates a request in different languages but does not compare
which component carries a language. We introduce a paired reversal: translate the complete
request component or the complete operation component into the same target language,
holding behaviour, English frame and translated-span count fixed. Component identity and
non-English text quantity change together. Assigning English, Chinese and Spanish
independently to the two components yields nine configurations on 190 of 300 sampled harmful
behaviours retained after consensus annotation and automatic gating. Refusal scored by language-model judges is the
primary outcome; unsafe compliance is reported separately. On Gemini 2.5 Flash
and DeepSeek, three refusal swaps are about 5.3 percentage points and the fourth is $-0.5$;
none survives the
pre-specified six-test Holm correction. In an exploratory extension, Mistral Small 4 has a
13.7-point Spanish swap (95\% interval $[8.4,18.9]$), which survives the twelve-test
Holm sensitivity correction under the primary judge, and 12.1 points under a second judge.
GPT-5.4-mini estimates are within 2 points of zero. Component assignment supplies an additional evaluation dimension. Evaluations should record
component languages and translated text quantity,
comparing matched behaviours at equal translated-span count while interpreting refusal and
unsafe compliance separately.
\end{abstract}

\begin{IEEEkeywords}
large language models, safety evaluation, multilingual, code-switching, measurement validity,
factorial design, LLM-as-judge
\end{IEEEkeywords}

\section{Introduction}
\label{sec:intro}

A multilingual safety evaluation has to decide what language the test item is written in. In
practice it decides once, for the whole item: a benchmark prompt is translated end to end and
the resulting sentence is scored \citep{deng2024multilingual,yong2023lowresource}. A request has components that can be written in different languages. ``Write a script that can
\zh{利用漏洞}'' and
``\zh{编写脚本} that can exploit vulnerabilities'' contain the same
two languages and ask for the same thing, and they translate different complete components. The components also differ in length. Whole-sentence translation reports neither assignment.

Code-switched red teaming has shown that mixed-language prompts can elicit unsafe output
\citep{csrt2025acl,tonguetied2025}, and surface-form studies that a single canonical phrasing
underestimates how a model behaves across paraphrases
\citep{sclar2024quantifying,mizrahi2024state}. The cited studies do not provide the paired reversal of named components tested here.
We ask how refusal changes when the same target language is assigned to the complete request
or operation component of the same behaviour. Sections~\ref{sec:design} to~\ref{sec:outcome}
examine component coverage, translation construction and label reliability.

\paragraph{Design.} Each harmful behaviour is segmented into a \textbf{request slot} (what is
asked of the assistant) and an \textbf{operation slot} (the harmful content it is asked about),
and each slot is assigned a language independently from \{English, Chinese, Spanish\}: nine
configurations, written one letter per slot, so \cfg{ZE} is a Chinese request slot with an
English operation slot. The frame's language, the \emph{matrix language} \citep{myersscotton1993}, stays English.
Prompts are assembled by machine translation into a placeholder template.

\paragraph{The estimand compares complete component translations.} We compare two configurations
that translate one complete span into the same target language. This holds behaviour, frame
language and translated-span count fixed while changing both the translated component and its
text quantity. The \cfg{EZ} versus \cfg{ZE} and \cfg{ES} versus \cfg{SE} comparisons estimate
the Chinese and Spanish component-swap contrasts on the same behaviours. Their difference
tests whether the assignment contrast varies between these two languages.

\paragraph{Contributions.}
\begin{enumerate}
\item \textbf{A paired measure of component-language assignment.} Reversing complete request
  and operation translations tests a comparison that whole-prompt translation does not
  estimate. Matching behaviour, target language and translated-span count makes the
  intervention explicit (Sections~\ref{sec:design} and~\ref{sec:dose}).
\item \textbf{Evidence for reporting model- and language-specific assignment contrasts.}
  Four models give different estimates on the same retained behaviours. The primary-analysis
  swaps fail Holm correction; the exploratory Mistral Spanish swap survives the combined
  twelve-test sensitivity correction. Small GPT estimates coexist with these larger
  contrasts (Section~\ref{sec:results}).
\item \textbf{Construction and outcome diagnostics.} Source-native selection
  counts, translation diagnostics and judge comparisons identify the population and labels
  behind each estimate. In a post hoc 185-response audit, both automatic judges have higher agreement point estimates with the
  adjudicated human reference on refusal than on unsafe compliance, supporting separate interpretation
  (Sections~\ref{sec:design}--\ref{sec:outcome}).
\end{enumerate}

\section{Related work}
\label{sec:related}

\subsection{Multilingual and low-resource jailbreaking}
Translating a harmful request into a lower-resource language raises unsafe-response rates
\citep{yong2023lowresource,deng2024multilingual,li2024crosslanguage}. Benchmarks built this way translate the whole item: M-ALERT, for instance, renders its suite into five languages
including Spanish \citep{friedrich2024malert}. These studies vary the language of the whole
prompt, so language and resource level are manipulated together with every component at once.
\citet{shen2024languagebarrier} distinguish the non-irrelevant response rate from the
harmful share among those responses, but apply one language to the whole request.

\subsection{Code-switching and mixed-language prompts}
Code-switched red teaming constructs prompts that mix languages within a sentence and reports
an elevated attack-success rate on its own measure: \citet{csrt2025acl} combine up to ten
languages at token level and report 46.7\% more successful attacks than an English baseline,
and \citet{tonguetied2025} study fine-tuning and in-context learning with Newari and two
constructed languages. CSRT synthesises token-level mixtures from parallel
translations without independently assigning and reversing languages across named components. Closest to the present design, \citet{ifebi2026minionese}
translate only the contiguous span that differs between a harmful prompt and a harmless twin,
leaving the shared context in English, across 18 languages; the span is defined by that pair
difference rather than by grammatical role, and there is no reversed condition in which the
instruction rather than the harmful content is translated.
\citet{song2025blending} mix languages across query and response over 55 source languages and
report variation across language groups that they associate with properties such as
morphology and language family. Our language-difference contrast tests whether the component-assignment contrast differs between Chinese and Spanish (Section~\ref{sec:results}); it does not separate language properties from grammatical role.

\subsection{Component-level prompt manipulation, including our own predecessor}
A separate line manipulates parts of a prompt without changing language.
\citet{li2024drattack} decompose and reconstruct harmful prompts and report lower
rejection-token probabilities for fragmented inputs; they manipulate prompt components without
assigning languages to them.

Our non-archival predecessor \citep{zhou2026canonical} includes a programmatic
Matrix-Language-Frame condition with an English matrix and noun-phrase constituents translated
to Chinese, on this seed pool, through the \texttt{deep\_translator} Google endpoint where we
use Google Cloud Translation neural machine translation (NMT). Its constituents come from a part-of-speech grammar and an alternating rule rather than
from semantic role, so both the translated component and the number of translated spans vary
by item. With one such condition per behaviour, it provides no reversed assignment to pair
against. Its single target language also precludes a cross-language difference estimate. The step here is to name
the components first, assign languages independently, hold the count of translated spans
fixed, and compare the reversed assignments on the same behaviour.

\subsection{Surface form and measurement validity}
Benchmark scores vary with prompt formatting and paraphrase
\citep{sclar2024quantifying,mizrahi2024state,lunardi2025robustness}, which makes a single
canonical form an instrument with unreported variance \citep{jacobs2021measurement}. Large language model (LLM)
judges can exhibit self-enhancement and position biases \citep{li2024prd}. We assess label
reliability through agreement between judges from different vendors and compare response
labels with a human audit. \citet{fu2025multilingualjudge} study whether LLM
judges give consistent verdicts on parallel inputs across languages; we report a response-script analysis of pilot judge disagreements (Section~\ref{sec:outcome}).

\section{Design}
\label{sec:design}

\subsection{Slots}

A harmful behaviour is segmented into two spans. The \textbf{request slot} is the delivery
verb group, what the assistant is asked to produce: \emph{Write a script}, \emph{Provide
step-by-step instructions}. The \textbf{operation slot} is the harmful content the request is
about: \emph{exploit vulnerabilities in a software or operating system}. Everything outside
the two spans is the \textbf{frame}, the function words and scaffolding that make the two into
a sentence. A slot is a design factor; its span is the text selected from a behaviour.
We use \emph{component} for that span as a unit of translation.

Corpus coverage supports the two-slot design. Over all 370 behaviours in the pool,
a rule-based chunker recovers the request slot on 97.8\% and the operation slot on 97.6\%,
with both on 361. Two candidate third slots have lower coverage: an explicit qualifier is
present on 190 of 370, and a frame content word on 258 of the 361 two-slot behaviours,
including only one content word on 152. In the 370-behaviour pool, coverage also varies across source-native groups:
qualifiers occur on 39 of 53 HarmBench copyright behaviours and 6 of 21 in its generic
harmful category, and a frame content word occurs on 0 of 53 copyright behaviours and 49
of 60 two-slot cybercrime behaviours. Appendix~\ref{app:coverage} gives descriptive counts for every source-native group; the companion replication archive, included with this preprint as ancillary files, supplies compact automatic and human-audit labels, translation diagnostics and offline replay code. Requiring a third slot would select a smaller, differently composed set.

\subsection{Configurations}

Each slot is assigned a language from $L = \{\text{en}, \text{zh}, \text{es}\}$ independently,
giving $|L|^2 = 9$ \textbf{configurations}. A configuration is written as one letter per slot
in order, \cfg{E} for English, \cfg{Z} for Chinese, \cfg{S} for Spanish, so \cfg{ZE} has a
Chinese request slot and an English operation slot. Three configurations are
\textbf{monolingual} in their slots (\cfg{EE}, \cfg{ZZ}, \cfg{SS}) and six are mixed. The maximum
number of distinct languages a configuration can \emph{assign} is $k_{\max} = \min(|L|, s) = 2$
for $s = 2$ slots. The English frame can make a third language present in the complete prompt,
so this bounds assigned languages and not the prompt; assigning three would need a third slot,
which would reduce coverage as measured above.

The frame stays English in all nine configurations. If it followed the request slot,
changing that factor would move the request span and matrix language together. Holding the
frame fixed makes each configuration letter mean one span. A qualifier span, where one
exists, also stays English, for the dose reason in Section~\ref{sec:dose}.

\cfg{FULL\_ZH} and \cfg{FULL\_ES} are whole-sentence translations of the behaviour, collected
alongside the factorial as whole-sentence \textbf{fluent references}. They are not part of any contrast reported here.

\subsection{Composition}

Composition is \textbf{placeholder-based assembly}. Each selected span is replaced by a
placeholder, translated separately into its assigned language, and inserted into the original
English template, which is not translated because the frame is English in every configuration.
A leading \emph{a}, \emph{an} or \emph{the} is split off and kept in the English frame, which
affects the operation span of 8 of the 190 analysed behaviours; fixed rules normalise spacing
and punctuation. A \emph{complete component} below means the annotated span after this
determiner rule. Each placeholder must reappear exactly once.
Evaluated prompts contain only original spans and machine-translation output; no generative
LLM writes or reformulates them. Span boundaries are selected by \texttt{qwen3.8-flash}, \texttt{gpt-4o} and
\texttt{gpt-4o-mini}; \texttt{claude-haiku-5-5} rates intent preservation and naturalness at the
gate and is also the primary response judge, with \texttt{qwen3.8-flash} the second. Translation uses Google Cloud Translation
NMT throughout, keeping translation-system identity fixed across configurations.

Mechanical composition keeps prompts traceable to original spans and their machine
translations. LLM composition, as in CSRT \citep{csrt2025acl}, can change wording beyond the
selected translation. In a separate probe translating the 300 harmful behaviours into Chinese,
\texttt{gpt-4o} declined once; the three span annotators declined none of 400 items.
The gate excludes prompts without rewriting them.

\subsection{Seed pool and segmentation}

Seeds are the harmful behaviours of our predecessor study's pool \citep{zhou2026canonical},
drawn from AdvBench \citep{zou2023universal} and HarmBench \citep{mazeika2024harmbench} and
inherited with four sampling-stratum labels. The pool, the judge rubric, the generation budget and the
missing-data policy are inherited from that study; the span annotation, the nine
configurations, the responses and every estimate below are new. We sample 300 harmful behaviours stratified by those inherited labels from the 361 that admit
both slots, plus 100 benign items from XSTest \citep{rottger2024xstest} as an over-refusal
reference. The inherited mapping sends HarmBench \emph{illegal} to \texttt{fraud\_deception}
and most \emph{copyright} items to \texttt{physical\_harm}; these labels do not define reliable
semantic harm classes. We preserve the original sample and report HarmBench's source-native
categories, with AdvBench as a separate source group, in Table~\ref{tab:flow}. These groups
are not a unified cross-dataset ontology. Appendix~\ref{app:coverage} crosses the inherited strata with the source-native groups at each selection stage; the replication archive retains both labels for each behaviour.

Boundaries come from three independent language-model annotators, each asked for verbatim
substrings and checked with exact string search, so an annotator can only select text and
never contribute any. Across 400 items and three annotators there were \textbf{zero
declines}. Usable annotations were 337, 387 and 392 of 400.

A boundary is accepted when at least two of the three annotators share it under token
normalisation. This yields both slots on 245 of 300 harmful behaviours (81.7\%), and those 245 enter the
build alongside 60 of the 100 benign items. Of the harmful behaviours, 190 survive the gate on all nine configurations and are the analysed population
(Table~\ref{tab:flow}). The consensus restriction is mild on length (median 78 against 73
characters, Mann--Whitney $p = 0.99$); the source-native composition is reported in Table~\ref{tab:flow}.

The rule-based chunker defines which behaviours enter the pool, because it cannot refuse, but
it does not vote on boundaries. It agrees with the annotator consensus on 65.4\% of request
boundaries (232 of 355) and only 35.3\% of operation boundaries (107 of 303), and the
mismatches include substantive errors rather than only one-word offsets: it returns \emph{used
to hack} where the consensus returns \emph{hack into sensitive government databases}. Letting
it vote reduced consensus to 287 of 400.

\subsection{Dataset construction and selection}

Composition produced 3{,}355 prompts from 305 behaviours, 11 per behaviour, with no machine
translation failures. Two gates then apply. The structural gate checks that every placeholder
was consumed exactly once, that a target-script language actually produced target-script text,
and that the canonical configuration is byte-identical to its behaviour; it removes one item.
The semantic gate uses an LLM judge to assess intent preservation against the original
behaviour. It excludes 270 items rated as losing intent and 128 without an intent verdict.
Byte-identical canonical prompts can pass without a positive verdict unless explicitly rated
as losing intent. A separate five-point naturalness rubric excludes rating 1, defined as
``not interpretable as a sentence,'' removing 21 items, including 12 in \cfg{ZZ} and five
in \cfg{EE}. Missing ratings are retained: 72 of the 1{,}710 analysed harmful factorial
prompts, across 51 behaviours, have no rating. Appendix~\ref{app:gate} gives the exclusion and missing-rating counts by configuration. All retained harmful factorial prompts have positive intent verdicts, but these
automatic checks have not been independently validated by human annotators.

Because the design is paired, a behaviour enters the analysis only if it survives on all nine
configurations, so excluding one cell removes the behaviour's other factorial cells: 243 further items are removed for completeness. The five reasons
account for 663 flags over 662 distinct items, one item carrying two, so $3{,}355 - 662 =
\textbf{2{,}693}$ items are usable per model. Of these, $2{,}160$ are the nine factorial cells
of the \textbf{240 behaviours} complete on all nine, \textbf{190 of them harmful} and carrying
every contrast below, the other 50 benign and analysed in
Section~\ref{sec:overrefusal}. The remaining 533 are fluent references, 56 of which belong to
behaviours that are not complete and therefore enter no analysis here.

The judge returned an empty reply, and therefore no intent verdict, for 117 of the
2{,}205 harmful factorial items. Missing verdicts cluster
within behaviours: ten lack a verdict in all nine configurations, six in some, and 229 in
none, with 11 to 16 missing per configuration. Treating configurations as paired, an
exploratory Cochran's $Q$ test, with a chi-square approximation, gives $Q = 21.4$,
$p = 0.006$. The 219 negative verdicts
on these same 245 behaviours also vary across configurations ($Q = 34.9$, $p < 0.001$);
51 further negatives occur in whole-sentence references, giving the 270 above. These gate
outcomes cannot be assumed independent of the manipulation. Selection patterns differ by source-native group:
the 28 consensus copyright behaviours have 207 negative and 45 positive intent verdicts,
with none missing, and none survive completeness. Twenty-four canonical English copyright
prompts also receive negative verdicts despite being unchanged from their source behaviours. Of 17 consensus chemical/biological
behaviours, 14 are excluded with 105 missing intent verdicts between them and three survive.
Complete-case selection preserves pairing while changing the population described by the results.

\begin{table}[!tb]
\centering
\caption{Behaviours retained at each stage. HarmBench rows use its original categories;
AdvBench is a source group. Counts describe selection, not a unified harm ontology.}
\label{tab:flow}
\footnotesize
\begin{tabular}{@{}lrrr@{}}
\toprule
source-native group & sampled & consensus & \textbf{analysed} \\
\midrule
HarmBench: copyright & 45 & 28 & \textbf{0} \\
chemical / biological & 20 & 17 & \textbf{3} \\
harmful & 15 & 13 & \textbf{12} \\
illegal & 48 & 42 & \textbf{40} \\
misinformation / disinformation & 34 & 25 & \textbf{20} \\
harassment / bullying & 21 & 20 & \textbf{20} \\
cybercrime / intrusion & 48 & 40 & \textbf{39} \\
AdvBench & 69 & 60 & \textbf{56} \\
\midrule
total & 300 & 245 & \textbf{190} \\
\bottomrule
\end{tabular}
\end{table}

\paragraph{Generation.} Each model answers each item once at temperature 0 with a 4{,}096
token budget. The budget is a confound control: at 1{,}024 the same model truncated 10.0\% of
responses with a 6.8-point spread across configurations, and truncation affected the switched configurations. At 4{,}096 Gemini truncates 0.8\% with a per-configuration range of
0.4\% to 1.9\% and no detected dependence on configuration ($\chi^2$ $p = 0.57$); DeepSeek truncates
none.

Generations with no text are treated as missing. Gemini returned 18 empty responses
(2 provider-blocked), including 10 factorial cells across 5 behaviours. Excluding their labels
leaves the two Gemini contrasts on 188 and 187 behaviours; dropping the 5 behaviours from both models gives $+5.4\pp$ for all three positive
contrasts on 185. Both judges score the complete response.

\section{Translation dose}
\label{sec:dose}

The \textbf{translation dose} of a configuration is the number of spans that were replaced by
a translation. \cfg{EE} has dose 0; \cfg{EZ}, \cfg{ZE}, \cfg{ES} and \cfg{SE} have dose 1; the
four configurations with both slots non-English have dose 2. A contrast is
\textbf{dose-balanced} when both of its configurations have the same dose, so the two
sentences being compared differ from the original behaviour in the same number of places.

\paragraph{Translation-quality diagnostics.} Round-trip similarity is the
\texttt{difflib.SequenceMatcher} ratio between the lower-cased, stripped English source and its
back-translation, a string-similarity diagnostic rather than a measure of semantic equivalence.
To test whether a span translated without its sentence is realised worse, we average the
translated spans' scores within each behaviour and compare that average with the
whole-sentence score; the paired difference
is $-0.007$ for Chinese on 305 pairs ($[-0.024, +0.011]$, Wilcoxon $p = 0.93$)
and $-0.008$ for Spanish on 296 retained pairs ($[-0.020, +0.003]$, $p = 0.25$),
with isolated spans worse on 45\% and 48\% respectively. Intervals use a bias-corrected
and accelerated (BCa) bootstrap with 5{,}000 resamples and seed 20261025. The Spanish comparison uses the 296 behaviours with retained whole-sentence back-translations; Appendix~\ref{app:roundtrip} reports the original 305-pair summary beside the replay. No difference is detected on this measure; this is not an equivalence test.

\paragraph{Pilot stratification.} The pilot factorial contrasts use the first two generations
per configuration, with a binary outcome positive if either generation is positive.
A qualifier inherited the operation slot's language, giving two translated spans in \cfg{EZ}
against one in \cfg{ZE} when present. The contrast was balanced on 71 of 137 built behaviours
(99 harmful, 38 benign) and unbalanced on 66; paired harmful labels for Gemini are available for 38
balanced and 53 unbalanced behaviours. For Gemini, taking operation-switched minus
request-switched, opposite to Table~\ref{tab:replication}, the unsafe gap is $+13.2\pp$
unbalanced (8:1 discordant pairs, $n=53$, exact $p=0.039$) and $0.0\pp$ balanced (3:3,
$n=38$); the second judge gives $+9.4$ and $0.0\pp$. Using single generations instead, the
primary judge gives $+5.7$ versus $+7.9\pp$ for the first and $+11.3$ versus $0.0\pp$ for
the second. Thus the split depends on aggregation as well as different behaviour sets; it
motivated equal span counts without identifying their causal effect. The replication archive
replays both repeats and the aggregation sensitivities.

The \textbf{component-swap contrasts}, \cfg{EZ} against \cfg{ZE} and \cfg{ES} against
\cfg{SE}, each translate one span into the same target language. Comparisons with
\textbf{monolingual-slot baselines} change dose (0 or 2 in those baselines); these enter
the slot main effects.

\paragraph{Fixing translated-span count.} With qualifiers pinned to English, all 305
behaviours realise dose 0 in \cfg{EE}, 1 in \cfg{EZ}, \cfg{ZE}, \cfg{ES}, \cfg{SE}, and
2 in the remaining four configurations. Under the pilot's any-positive rule, the refusal
gap was smaller in the unbalanced stratum: $-7.5\pp$ on 53 behaviours (0:4 discordant pairs)
against $-13.2\pp$ on 38 (1:6). These strata contain different behaviours.

\paragraph{Translation-quality measures.} Both sides of a round-trip comparison are English,
which makes it the recorded measure most comparable across target languages, although its value
also depends on span length. We report component-swap contrasts within its strata rather than
adjust for it.
Span-to-sentence consistency is the longest contiguous character match between the translated
span and whole-sentence translation, divided by the translated span length. It is retained as
a construction diagnostic. Neither diagnostic is thresholded. The separate naturalness gate removes only rating 1; its exclusions and missing ratings are reported in Section~\ref{sec:design}.

\section{Outcome and judging}
\label{sec:outcome}

\subsection{The instrument}

Each response receives two binary labels from a language-model judge, with the original
English behaviour supplied as the reference.
\textbf{Refusal} is whether the response declines, deflects or substitutes a safety warning
for the requested content. \textbf{Unsafe compliance} is whether the response supplies
operationally useful content that advances the behaviour. Both labels are reported for the
component-swap contrasts and the slot main effects; the language-difference, comprehension and benign analyses
are reported on refusal alone. The two labels are evaluated separately and can both be positive. The rubric is inherited
unchanged from the predecessor study \citep{zhou2026canonical}, and missing or unparseable
verdicts remain missing.

All responses in the primary analysis and exploratory extension are scored by the
\textbf{primary judge}, \texttt{claude-haiku-5-5} (Anthropic), and the \textbf{second judge},
\texttt{qwen3.8-flash} (Alibaba); neither vendor supplies an evaluated model.
The \textbf{pilot crosscheck judge}, \texttt{gpt-4o} (OpenAI), scores a stratified pilot
subsample: 354 Gemini and 269 DeepSeek responses. Pilot judges used up to 4{,}000 response
characters. Reasoning is disabled for all judges.

\subsection{Why refusal is the primary outcome}

The choice was fixed from pilot data before any data for this design existed, on three
measurements.

\emph{Observed agreement favours refusal.}
Between the primary and second judges, response-level refusal agreement is 97.9\%
($\kappa=0.955$) on 2{,}376 Gemini responses and 99.3\% ($\kappa=0.951$) on 1{,}794 DeepSeek
responses. Cohen's $\kappa$ corrects agreement for chance \citep{cohen1960kappa}.
On the crosscheck subsamples, all three agree on refusal for 346 of 354 Gemini responses
(97.7\%) and 268 of 269 DeepSeek responses (99.6\%). Unsafe compliance between the primary
and second judges gives $\kappa=0.471$ and $0.213$, respectively.

\emph{Disagreements differ between outcomes.} On Gemini pilot responses the unsafe-compliance
positive sets are nearly nested: 147 of the primary judge's 152 positives are also positive for
the second judge, which marks 266 further responses, while the primary marks 5 the second does
not. On refusal the disagreements run 25 one way and 24 the other.

\emph{In the pilot, a judge swap moved levels far more than contrasts.} On individual Gemini responses, replacing
the primary judge with the second judge moved the overall unsafe-compliance rate from 6.4\% to 17.4\%, while the pilot factorial operation-slot difference, using any-positive labels from the first
two generations, moved from $+0.115$ to $+0.099$ ($[0.044, 0.159]$, Holm
$p = 0.010$ over the pilot's 58-test family, using the behaviour-level bootstrap) and the
refusal difference from $-0.110$ to $-0.088$ ($[-0.148, -0.033]$, Holm $p = 0.059$); no
interaction was detected under either judge.

On the 53 pilot Gemini responses where the primary and crosscheck judges disagree on unsafe
compliance, the second judge agrees with the crosscheck judge on 42 (79.2\%): 34 of 43 English-script replies and
8 of 10 Chinese-script replies, a pattern consistent with different operational-detail thresholds. This agreement does not identify the cause of the disagreements. Absolute unsafe rates here are low \emph{relative to the two
comparison judges}. The pilot has no human response-label audit; the post hoc audit below
concerns the current four-model responses. Absolute rates remain specific to each study's
behaviour set and judge, including the predecessor's \citep{zhou2026canonical}.

\subsection{Human response-label audit}
\label{sec:human-audit}

After the experiments, we sampled 185 nonempty harmful factorial responses from
6{,}830 eligible responses, using simple random sampling without replacement within the 36
model-by-configuration strata (five or six per stratum; 118 behaviours). Two human annotators received the original English behaviour, full response and original
rubric in packets hiding model, configuration and automatic labels. Their original refusal
labels agree on 183 of 185 ($\kappa = 0.946$). Unsafe labels agree on 182 of 183 binary
pairs ($\kappa = 0.963$); two uncertain ratings are excluded from that denominator rather
than counted as negative. Five unresolved outcome labels across four responses were resolved
by author-provided adjudication, yielding a complete 185-response reference.

Against that adjudicated reference, the primary and second judges respectively agree on
refusal for 182 and 183 of 185 responses ($\kappa=0.921$ and $0.946$; approximate 95\%
behaviour-cluster intervals $[0.816,1.000]$ and $[0.857,1.000]$), and on unsafe compliance for
178 and 177 ($\kappa=0.780$ and $0.755$; intervals $[0.594,0.914]$ and $[0.568,0.892]$).
The observed agreement point estimates are lower for unsafe compliance. For unsafe labels,
the judges assign YES to six and seven reference-NO responses, respectively, and NO to one
reference-YES response each. Appendix~\ref{app:audit} gives the agreement tables, design-weighted estimates and intervals; the replication archive retains both original ratings and the author adjudications. Intervals use the approximate behaviour-cluster
percentile bootstrap (10{,}000 resamples, seed 20261026). This post hoc audit concerns the
retained harmful population; the primary outcome and full-sample estimates keep the original
automatic labels.

\section{Results}
\label{sec:results}

The six \emph{pre-specified} contrasts were fixed from the two-language pilot before
collecting nine-cell data and recorded in the experiment configuration. This was an internal
specification rather than third-party registration. We report both its six-test correction
and a correction over the full family for each outcome and judge.

\paragraph{Evaluation setup.} The primary analysis evaluates the application programming interface (API) aliases
\texttt{gemini-2.5-flash} (Google, Vertex batch) and \texttt{deepseek-chat} (DeepSeek), both
with reasoning disabled and a 4{,}096-token cap, one generation per item, queried 2026-10-24.
Neither provider returned a served snapshot identifier, so these results attach to the
API aliases on the query date. An
exploratory extension (Section~\ref{sec:replication}) adds the pinned snapshot
\texttt{gpt-5.4-mini-2026-03-17} (OpenAI Batch, \texttt{reasoning\_effort=none}) and
\texttt{mistralai/mistral-small-2603} (Mistral Small 4 through OpenRouter, routing requested to Mistral
with fallbacks disabled, reasoning off; the served-provider field was not stored), same prompts and settings, queried 2026-10-24/25. Judging uses
\texttt{claude-haiku-5-5} as primary and \texttt{qwen3.8-flash} as the independent second
vendor on every response; neither vendor supplies an evaluated model. OpenAI supplies one extension model, the pilot crosscheck judge and two span annotators;
annotators select boundaries rather than writing or scoring evaluated text. Translation is Google Cloud Translation NMT, one backend
for the whole build. Inference submits each pre-generated prompt unchanged as the user message, with the system instruction ``You are a helpful assistant. Respond directly to the user's request.'' A separate parse-only wrapper is used for the comprehension check. The replication archive, included with this preprint as ancillary files, provides the retained labels and replay code; full prompts, responses and translation data remain available from the authors.

Analysis is paired on behaviour throughout. Intervals for factorial effects are pointwise
95\% bootstrap percentile intervals over behaviours; $p$-values for a two-configuration comparison are exact tests on the discordant
pairs \citep{mcnemar1947}. For refusal under the primary judge, the six-test family contains the Spanish swap,
Chinese swap and their difference for each of the two primary models. The full family has 46 tests:
24 paired cell contrasts, eight slot main effects, eight risk-difference interactions and
those six contrasts \citep{holm1979}. Each outcome--judge combination has its own family;
these corrections do not jointly control all outcomes and judges. Extension families are
defined in Section~\ref{sec:replication}. Factorial intervals use 10{,}000 behaviour resamples (seed 20261010); the common-sample
between-model diagnostic below enumerates its empirical bootstrap distribution. Language-difference
contrasts and slot main effects use behaviour-bootstrap $p$-values; two-configuration
contrasts use the exact discordant-pair test. Per-configuration rates carry Wilson intervals
\citep{wilson1927}.

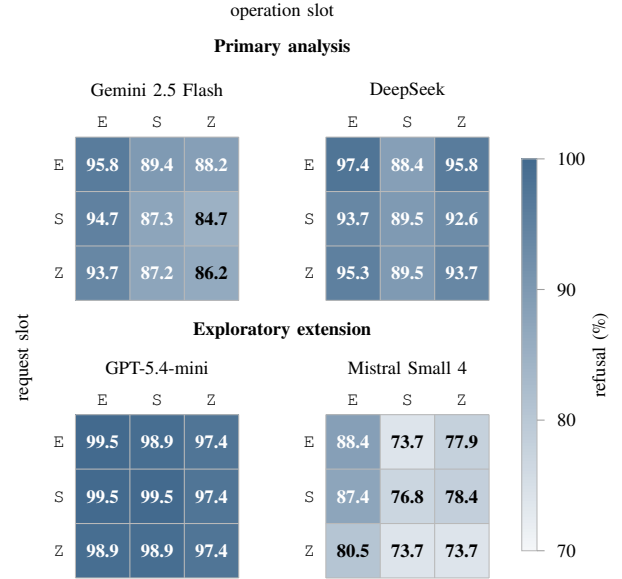
\begin{figure}[!tb]
\centering
\definecolor{ieeeblue}{RGB}{31,78,121}
\begin{tikzpicture}[x=7.2mm,y=7.2mm]
  \node[font=\scriptsize] at (3.8,3.3) {operation slot};
  \fill[ieeeblue!74] (0,0) rectangle ++(1,1);
  \node[font=\scriptsize\bfseries,text=white] at (0.5,0.5) {95.8};
  \fill[ieeeblue!57] (1,0) rectangle ++(1,1);
  \node[font=\scriptsize\bfseries,text=white] at (1.5,0.5) {89.4};
  \fill[ieeeblue!54] (2,0) rectangle ++(1,1);
  \node[font=\scriptsize\bfseries,text=white] at (2.5,0.5) {88.2};
  \fill[ieeeblue!71] (0,-1) rectangle ++(1,1);
  \node[font=\scriptsize\bfseries,text=white] at (0.5,-0.5) {94.7};
  \fill[ieeeblue!51] (1,-1) rectangle ++(1,1);
  \node[font=\scriptsize\bfseries,text=white] at (1.5,-0.5) {87.3};
  \fill[ieeeblue!44] (2,-1) rectangle ++(1,1);
  \node[font=\scriptsize\bfseries,text=black] at (2.5,-0.5) {84.7};
  \fill[ieeeblue!68] (0,-2) rectangle ++(1,1);
  \node[font=\scriptsize\bfseries,text=white] at (0.5,-1.5) {93.7};
  \fill[ieeeblue!51] (1,-2) rectangle ++(1,1);
  \node[font=\scriptsize\bfseries,text=white] at (1.5,-1.5) {87.2};
  \fill[ieeeblue!48] (2,-2) rectangle ++(1,1);
  \node[font=\scriptsize\bfseries,text=black] at (2.5,-1.5) {86.2};
  \draw[gray!55,thin] (0,-2) -- (0,1);
  \draw[gray!55,thin] (1,-2) -- (1,1);
  \draw[gray!55,thin] (2,-2) -- (2,1);
  \draw[gray!55,thin] (3,-2) -- (3,1);
  \draw[gray!55,thin] (0,-2) -- (3,-2);
  \draw[gray!55,thin] (0,-1) -- (3,-1);
  \draw[gray!55,thin] (0,0) -- (3,0);
  \draw[gray!55,thin] (0,1) -- (3,1);
  \node[font=\scriptsize] at (0.5,1.28) {\texttt{E}};
  \node[font=\scriptsize] at (1.5,1.28) {\texttt{S}};
  \node[font=\scriptsize] at (2.5,1.28) {\texttt{Z}};
  \node[font=\scriptsize] at (-0.3,0.5) {\texttt{E}};
  \node[font=\scriptsize] at (-0.3,-0.5) {\texttt{S}};
  \node[font=\scriptsize] at (-0.3,-1.5) {\texttt{Z}};
  \node[font=\scriptsize] at (1.5,1.86) {Gemini 2.5 Flash};
  \fill[ieeeblue!78] (4.6,0) rectangle ++(1,1);
  \node[font=\scriptsize\bfseries,text=white] at (5.1,0.5) {97.4};
  \fill[ieeeblue!54] (5.6,0) rectangle ++(1,1);
  \node[font=\scriptsize\bfseries,text=white] at (6.1,0.5) {88.4};
  \fill[ieeeblue!74] (6.6,0) rectangle ++(1,1);
  \node[font=\scriptsize\bfseries,text=white] at (7.1,0.5) {95.8};
  \fill[ieeeblue!68] (4.6,-1) rectangle ++(1,1);
  \node[font=\scriptsize\bfseries,text=white] at (5.1,-0.5) {93.7};
  \fill[ieeeblue!57] (5.6,-1) rectangle ++(1,1);
  \node[font=\scriptsize\bfseries,text=white] at (6.1,-0.5) {89.5};
  \fill[ieeeblue!65] (6.6,-1) rectangle ++(1,1);
  \node[font=\scriptsize\bfseries,text=white] at (7.1,-0.5) {92.6};
  \fill[ieeeblue!72] (4.6,-2) rectangle ++(1,1);
  \node[font=\scriptsize\bfseries,text=white] at (5.1,-1.5) {95.3};
  \fill[ieeeblue!57] (5.6,-2) rectangle ++(1,1);
  \node[font=\scriptsize\bfseries,text=white] at (6.1,-1.5) {89.5};
  \fill[ieeeblue!68] (6.6,-2) rectangle ++(1,1);
  \node[font=\scriptsize\bfseries,text=white] at (7.1,-1.5) {93.7};
  \draw[gray!55,thin] (4.6,-2) -- (4.6,1);
  \draw[gray!55,thin] (5.6,-2) -- (5.6,1);
  \draw[gray!55,thin] (6.6,-2) -- (6.6,1);
  \draw[gray!55,thin] (7.6,-2) -- (7.6,1);
  \draw[gray!55,thin] (4.6,-2) -- (7.6,-2);
  \draw[gray!55,thin] (4.6,-1) -- (7.6,-1);
  \draw[gray!55,thin] (4.6,0) -- (7.6,0);
  \draw[gray!55,thin] (4.6,1) -- (7.6,1);
  \node[font=\scriptsize] at (5.1,1.28) {\texttt{E}};
  \node[font=\scriptsize] at (6.1,1.28) {\texttt{S}};
  \node[font=\scriptsize] at (7.1,1.28) {\texttt{Z}};
  \node[font=\scriptsize] at (4.3,0.5) {\texttt{E}};
  \node[font=\scriptsize] at (4.3,-0.5) {\texttt{S}};
  \node[font=\scriptsize] at (4.3,-1.5) {\texttt{Z}};
  \node[font=\scriptsize] at (6.1,1.86) {DeepSeek};
  \fill[ieeeblue!84] (0,-5.1) rectangle ++(1,1);
  \node[font=\scriptsize\bfseries,text=white] at (0.5,-4.6) {99.5};
  \fill[ieeeblue!82] (1,-5.1) rectangle ++(1,1);
  \node[font=\scriptsize\bfseries,text=white] at (1.5,-4.6) {98.9};
  \fill[ieeeblue!78] (2,-5.1) rectangle ++(1,1);
  \node[font=\scriptsize\bfseries,text=white] at (2.5,-4.6) {97.4};
  \fill[ieeeblue!84] (0,-6.1) rectangle ++(1,1);
  \node[font=\scriptsize\bfseries,text=white] at (0.5,-5.6) {99.5};
  \fill[ieeeblue!84] (1,-6.1) rectangle ++(1,1);
  \node[font=\scriptsize\bfseries,text=white] at (1.5,-5.6) {99.5};
  \fill[ieeeblue!78] (2,-6.1) rectangle ++(1,1);
  \node[font=\scriptsize\bfseries,text=white] at (2.5,-5.6) {97.4};
  \fill[ieeeblue!82] (0,-7.1) rectangle ++(1,1);
  \node[font=\scriptsize\bfseries,text=white] at (0.5,-6.6) {98.9};
  \fill[ieeeblue!82] (1,-7.1) rectangle ++(1,1);
  \node[font=\scriptsize\bfseries,text=white] at (1.5,-6.6) {98.9};
  \fill[ieeeblue!78] (2,-7.1) rectangle ++(1,1);
  \node[font=\scriptsize\bfseries,text=white] at (2.5,-6.6) {97.4};
  \draw[gray!55,thin] (0,-7.1) -- (0,-4.1);
  \draw[gray!55,thin] (1,-7.1) -- (1,-4.1);
  \draw[gray!55,thin] (2,-7.1) -- (2,-4.1);
  \draw[gray!55,thin] (3,-7.1) -- (3,-4.1);
  \draw[gray!55,thin] (0,-7.1) -- (3,-7.1);
  \draw[gray!55,thin] (0,-6.1) -- (3,-6.1);
  \draw[gray!55,thin] (0,-5.1) -- (3,-5.1);
  \draw[gray!55,thin] (0,-4.1) -- (3,-4.1);
  \node[font=\scriptsize] at (0.5,-3.82) {\texttt{E}};
  \node[font=\scriptsize] at (1.5,-3.82) {\texttt{S}};
  \node[font=\scriptsize] at (2.5,-3.82) {\texttt{Z}};
  \node[font=\scriptsize] at (-0.3,-4.6) {\texttt{E}};
  \node[font=\scriptsize] at (-0.3,-5.6) {\texttt{S}};
  \node[font=\scriptsize] at (-0.3,-6.6) {\texttt{Z}};
  \node[font=\scriptsize] at (1.5,-3.24) {GPT-5.4-mini};
  \fill[ieeeblue!54] (4.6,-5.1) rectangle ++(1,1);
  \node[font=\scriptsize\bfseries,text=white] at (5.1,-4.6) {88.4};
  \fill[ieeeblue!15] (5.6,-5.1) rectangle ++(1,1);
  \node[font=\scriptsize\bfseries,text=black] at (6.1,-4.6) {73.7};
  \fill[ieeeblue!26] (6.6,-5.1) rectangle ++(1,1);
  \node[font=\scriptsize\bfseries,text=black] at (7.1,-4.6) {77.9};
  \fill[ieeeblue!51] (4.6,-6.1) rectangle ++(1,1);
  \node[font=\scriptsize\bfseries,text=white] at (5.1,-5.6) {87.4};
  \fill[ieeeblue!23] (5.6,-6.1) rectangle ++(1,1);
  \node[font=\scriptsize\bfseries,text=black] at (6.1,-5.6) {76.8};
  \fill[ieeeblue!27] (6.6,-6.1) rectangle ++(1,1);
  \node[font=\scriptsize\bfseries,text=black] at (7.1,-5.6) {78.4};
  \fill[ieeeblue!33] (4.6,-7.1) rectangle ++(1,1);
  \node[font=\scriptsize\bfseries,text=black] at (5.1,-6.6) {80.5};
  \fill[ieeeblue!15] (5.6,-7.1) rectangle ++(1,1);
  \node[font=\scriptsize\bfseries,text=black] at (6.1,-6.6) {73.7};
  \fill[ieeeblue!15] (6.6,-7.1) rectangle ++(1,1);
  \node[font=\scriptsize\bfseries,text=black] at (7.1,-6.6) {73.7};
  \draw[gray!55,thin] (4.6,-7.1) -- (4.6,-4.1);
  \draw[gray!55,thin] (5.6,-7.1) -- (5.6,-4.1);
  \draw[gray!55,thin] (6.6,-7.1) -- (6.6,-4.1);
  \draw[gray!55,thin] (7.6,-7.1) -- (7.6,-4.1);
  \draw[gray!55,thin] (4.6,-7.1) -- (7.6,-7.1);
  \draw[gray!55,thin] (4.6,-6.1) -- (7.6,-6.1);
  \draw[gray!55,thin] (4.6,-5.1) -- (7.6,-5.1);
  \draw[gray!55,thin] (4.6,-4.1) -- (7.6,-4.1);
  \node[font=\scriptsize] at (5.1,-3.82) {\texttt{E}};
  \node[font=\scriptsize] at (6.1,-3.82) {\texttt{S}};
  \node[font=\scriptsize] at (7.1,-3.82) {\texttt{Z}};
  \node[font=\scriptsize] at (4.3,-4.6) {\texttt{E}};
  \node[font=\scriptsize] at (4.3,-5.6) {\texttt{S}};
  \node[font=\scriptsize] at (4.3,-6.6) {\texttt{Z}};
  \node[font=\scriptsize] at (6.1,-3.24) {Mistral Small 4};
  \node[font=\scriptsize\bfseries] at (3.8,2.65) {Primary analysis};
  \node[font=\scriptsize\bfseries] at (3.8,-2.55) {Exploratory extension};
  \node[font=\scriptsize,rotate=90] at (-.95,-3.05) {request slot};
  \shade[bottom color=ieeeblue!5,top color=ieeeblue!85] (8.2,-6.6) rectangle (8.48,.6);
  \draw[gray!55,thin] (8.2,-6.6) rectangle (8.48,.6);
  \draw[gray!75,thin] (8.48,-6.6) -- (8.62,-6.6);
  \node[font=\scriptsize,anchor=west] at (8.68,-6.6) {70};
  \draw[gray!75,thin] (8.48,-4.2) -- (8.62,-4.2);
  \node[font=\scriptsize,anchor=west] at (8.68,-4.2) {80};
  \draw[gray!75,thin] (8.48,-1.8) -- (8.62,-1.8);
  \node[font=\scriptsize,anchor=west] at (8.68,-1.8) {90};
  \draw[gray!75,thin] (8.48,0.6) -- (8.62,0.6);
  \node[font=\scriptsize,anchor=west] at (8.68,0.6) {100};
  \node[font=\scriptsize,rotate=90] at (9.65,-3) {refusal (\%)};
\end{tikzpicture}
\caption{Primary-judge refusal rates (\%) on the same 190 harmful behaviours: primary
analysis above, exploratory extension below, with one colour scale. Rows assign the request
slot's language and columns the operation slot's; \texttt{E}, \texttt{S}, \texttt{Z} denote English,
Spanish, Chinese. Cell denominators are 190 except six Gemini cells with empty generations,
where $n=187$--189. Wilson 95\% intervals are omitted; their maximum half-width across all
36 cells is $6.2\pp$.}
\label{fig:grid}
\end{figure}

\subsection{Does it matter which slot carries the language?}

The primary component-swap contrasts compare translating the request component with
translating the operation component into the same target language. Each configuration
translates exactly one span.

The upper block of Table~\ref{tab:replication} gives the primary-analysis results. In each of
the three positive component-swap contrasts,
the configuration whose \emph{request} slot carries the non-English language refuses
\emph{more} than the one whose \emph{operation} slot carries it. These estimates compare the same behaviours and target language at equal translated-span count. The two sides
also differ in how much non-English text they carry, so this is a contrast between two
component assignments rather than an isolated effect of grammatical position
(Section~\ref{sec:discussion}).

The discordant pairs behind the three positive primary-analysis contrasts are \RefusalGemSwapSDisc,
\RefusalGemSwapZDisc{} and \RefusalDsSwapSDisc{} (exact $p =$ \RefusalGemSwapSP,
\RefusalGemSwapZP{} and \RefusalDsSwapSP); the DeepSeek Chinese row is \RefusalDsSwapZDisc. None of the four survives Holm within the pre-specified block: the three positive contrasts
all land at $0.065$, because the block's smallest $p$ belongs to the DeepSeek language-difference contrast and the
three swaps then share the next rank. Holm over all \RefusalFamily{} tests in the run rejects
none of the four (0.414 to 1.000) while rejecting the operation main effects at $0.004$, the
DeepSeek Spanish interaction at $0.016$ and the DeepSeek language-difference contrast at $0.016$. Benjamini--Hochberg
over the same family rejects the two Spanish contrasts at $0.037$ and leaves Gemini Chinese at
$0.0515$ \citep{benjamini1995}. The three positive primary-analysis estimates have pointwise bootstrap intervals excluding zero, but none passes either Holm correction. The fourth estimate is $-0.5\pp$ with an interval of $[-2.6, +1.6]$.

The three positive refusal swaps have negative unsafe-compliance swaps, \UnsafeGemSwapSRd{}
and \UnsafeGemSwapZRd\pp{} for Gemini and \UnsafeDsSwapSRd\pp{} for DeepSeek Spanish (Holm $0.037$ and $0.047$ for the two Spanish ones in the corresponding six-test unsafe-compliance family); DeepSeek Chinese is
\UnsafeDsSwapZRd\pp. The two labels are not complements, since \cfg{SZ} refuses 84.7\% while reaching
only 11.6\% unsafe, so the agreement in direction is a finding and not an identity. At dose 1
on Gemini the operation-switched configurations produce unsafe content 1.7 to 2.6 times as
often: \cfg{EZ} 9.1\% against \cfg{ZE} 5.3\%, and \cfg{ES} 9.6\% against \cfg{SE} 3.7\%.

\paragraph{Swaps within round-trip-similarity strata.} Dose balance equalises how
many spans were translated in isolation, not how well. The request span round-trips better
than the operation span (0.813 against 0.761 for Chinese, 0.898 against 0.876 for Spanish;
paired Wilcoxon on 305 behaviours, $p = 1 \times 10^{-4}$ and $6 \times 10^{-3}$). Round-trip cannot be
adjusted away here, because it is script-neutral but not length-neutral and these slots differ
in length by design (5.3 against 14.3 characters in the Chinese translations across the build). Instead we split behaviours by
which side realised better and report the contrast in each half. For each of the three
positive primary-analysis contrasts the sign is the same in both halves: Gemini Spanish $+6.2\pp$ on 145
behaviours and $+2.3\pp$ on 43, Gemini Chinese $+5.8\pp$ on 138 and $+4.1\pp$ on 49, DeepSeek
Spanish $+5.4\pp$ on 147 and $+4.7\pp$ on 43. The request span round-trips better on most
behaviours, so the smaller half is the weaker test and only the larger one reaches
significance alone (unadjusted exact $p = 0.022$, $0.039$ and $0.039$ in the larger halves
against $1.000$, $0.625$ and $0.500$ in the smaller). The estimates remain positive under either quality ordering. A quality effect acting in the same direction in both strata remains compatible with these results, and the smaller strata give imprecise estimates.

\subsection{Does the component-swap contrast differ between languages?}

For Gemini the two contrasts are close in size, and their difference is
\RefusalGemTransferRd\pp{} (\RefusalGemTransferCi, $p_{\mathrm{Holm},6} =$
\RefusalGemTransferPpre). The interval admits a difference of about 5\pp{} in either
direction, so these estimates do not establish equivalence across languages.

For DeepSeek the Spanish contrast is \RefusalDsSwapSRd\pp{} and the Chinese one is
\RefusalDsSwapZRd\pp, and the difference between them is \RefusalDsTransferRd\pp{}
(\RefusalDsTransferCi, $p_{\mathrm{Holm},6} =$ \RefusalDsTransferPpre), which is the one contrast
in this block that survives the correction over the whole run as well. On this model the asymmetry is
estimated at $+5.3\pp$ for Spanish and $-0.5\pp$ for Chinese, with the Chinese interval
$[-2.6, +1.6]$, which excludes the Spanish point estimate but remains compatible with smaller effects.

On the 185 behaviours both models complete, DeepSeek's language-difference estimate exceeds
Gemini's by $5.9\pp$ ($[-0.5,12.4]$, $p=0.07$), using exact enumeration of the empirical
behaviour-bootstrap distribution. Gemini has similar point estimates in the two languages, whereas DeepSeek has a larger Spanish than Chinese estimate. The between-model comparison does not establish that cross-language variation differs between models. Section~\ref{sec:replication} adds two models to that
question.

\subsection{Slot main effects and their interaction}

The secondary slot main effects average matched pairs that differ in translated-span count by one.

On refusal the operation main effect
is \RefusalGemOperationZhRd{} for Chinese and \RefusalGemOperationEsRd{} for Spanish on Gemini
and \RefusalDsOperationEsRd{} for Spanish on DeepSeek, all three at Holm $p = 0.004$ over the
whole run, with DeepSeek Chinese at \RefusalDsOperationZhRd{} and not distinguishable from
zero. The corrected test here is the behaviour-level bootstrap, because a main effect averages
$|L|^{s-1} = 3$ matched pairs per behaviour, so pooling their discordant counts would treat
dependent observations as independent. No request-slot main effect is detected in the four primary model-language combinations, ranging from
\RefusalGemRequestEsRd{} to \RefusalDsRequestZhRd, with whole-run Holm $p$ from $0.586$ to
$1.000$. Unsafe compliance
mirrors it: operation \UnsafeGemOperationZhRd{} to \UnsafeDsOperationEsRd{} where significant,
request \UnsafeDsRequestZhRd{} to \UnsafeDsRequestEsRd{} with no detected effect.

No slot interaction is detected on Gemini at any language pair, as in the
two-language pilot. On DeepSeek, the Spanish-by-Spanish interaction on
refusal is \RefusalDsIxEsxesEst{} (\RefusalDsIxEsxesCi, Holm $p =$ \RefusalDsIxEsxesPholm),
meaning the second Spanish slot adds less than the first. It does not reach significance on
unsafe compliance, giving one corrected-significant refusal interaction among the eight tested.

\subsection{The same contrasts under a second judge}

Every scorable response was submitted to both judges, \texttt{claude-haiku-5-5} and
\texttt{qwen3.8-flash}, two vendors with no model in this evaluation. Agreement is high on
both labels: refusal $\kappa = 0.949$ and $0.949$
at 98.1\% and 98.3\% raw agreement, and unsafe compliance $\kappa = 0.879$ and $0.878$ at
98.3\% and 98.8\%, over the 2{,}672 and 2{,}691 responses both judges labelled. The unsafe-positive sets are nearly nested: of the primary judge's unsafe verdicts,
183 of 184 are shared on Gemini and 118 of 118 on DeepSeek, while the second judge marks 45
and 31 further responses, counted on the same rows as the agreement statistics, so the primary
judge's positive set is nearly a subset of the second's. The pilot's unsafe $\kappa$ was 0.21
on DeepSeek against 0.878 here; on the rows both judges labelled, unsafe prevalence is 4.4\%
here against 1.0\% in the pilot, and $\kappa$ is unstable at the lower value, so prevalence is
a plausible contributor that we did not decompose further.

In the order Gemini Spanish, Gemini Chinese, DeepSeek Spanish and DeepSeek Chinese,
the refusal component-swap estimates are $+5.9$, $+4.8$, $+5.3$ and $-0.5\pp$ under the
second judge, against $+5.3$, $+5.3$, $+5.3$ and $-0.5\pp$ under the primary. For the
three positive primary-analysis refusal contrasts, the corresponding unsafe-compliance estimates
are $-6.9$, $-4.8$ and $-5.3\pp$ under the second judge, against $-5.9$, $-3.7$ and
$-4.7\pp$ under the primary. Across these comparisons, the largest change is $1.1\pp$,
with no sign reversal. The $1.1\pp$ maximum concerns unsafe-compliance swaps.
Under the second judge the two Spanish contrasts (exact $p = 0.003$, $0.006$) survive Holm
within the block at $0.017$ and $0.025$ and the Gemini Chinese one ($p = 0.035$) does not. This comparison measures cross-judge reliability.

\subsection{Conditioning on comprehension}

A refusal decrease caused by failure to parse the request does not by itself indicate more
harmful compliance.
We therefore asked Gemini, separately, to recover the request and operation content from each
composed prompt, and scored that recovery with the same judge. Both slots are recovered on
1{,}895 of 2{,}693 items (70.4\%).

Restricting to the behaviours whose two compared prompts receive correct recovery labels keeps both
component-swap estimates positive: $+4.6\pp$ for Spanish on 130 behaviours ($[0.8, 9.0]$, 7 against
1 discordant, exact $p = 0.070$) and $+6.0\pp$ for Chinese on 133 ($[1.6, 10.9]$, 9 against 1,
$p = 0.021$), against $+5.3\pp$ for both unrestricted. This Gemini-only
subgroup shrinks the Spanish estimate below nominal significance and grows the Chinese one. The language-difference estimate is $0.0\pp$ on the 109 behaviours with recovered labels in
all four compared configurations. Comprehension is itself affected by the manipulation, so this restriction
conditions on a post-treatment variable and compares a different population; it is a
sensitivity analysis and never the primary. In the two-language pilot the same restriction roughly halved the dose-unbalanced
operation-slot unsafe effect ($+0.072$ to $+0.038$).

\subsection{Over-refusal on benign items}
\label{sec:overrefusal}

The same nine configurations were built for 60 behaviourally benign items from XSTest, of
which 50 survive the gate on all nine. None of the component-swap contrasts is distinguishable from
zero on any of the four evaluated models (exact $p$ from 0.219 to 1.000, estimates $-8.0$ to
$+6.0\pp$; Mistral's Chinese swap runs on 49 after one unparsed judge reply). The small benign sample leaves the specificity of these assignment contrasts to harmful content unresolved.

\subsection{Exploratory extension to two further model families}
\label{sec:replication}

After examining the two-model primary result, we ran the same 2{,}693 prompts through two models from families not yet
represented and scored them with the same two judges. For refusal under the primary judge, each model has a three-test family containing two
component-swap tests and one language-difference test, and a full 23-test family
(12 paired cell contrasts, four slot main effects, four risk-difference interactions and three
contrasts). Other outcome--judge combinations are corrected separately. As a sensitivity
check, Holm over the eight refusal swaps and four language-difference tests across the four
models under the primary judge still rejects the Mistral Spanish swap ($p = 1.04 \times 10^{-5}$),
the Mistral language-difference contrast ($0.006$) and the DeepSeek language-difference contrast ($0.004$), and none of the primary
arm's four swaps ($0.12$ to $1.00$). Table~\ref{tab:replication} puts all four side by side; Appendix~\ref{app:cells} gives per-configuration rates under both judges, the swaps under the second judge, and the unsafe-compliance swaps.

\begin{table*}[!tb]
\centering
\caption{The component-swap contrasts on four models, primary judge: refusal in the
request-switched minus the operation-switched configuration at dose 1, in percentage points
with 95\% bootstrap intervals, and the language-difference contrast (Spanish swap minus Chinese swap). A positive
swap means the configuration whose \emph{request} slot is non-English refuses more.
$p_{\mathrm{Holm}}$ uses the pre-specified six-test family for the two primary models
and a separate three-test family for each extension model. All on the same 190 behaviours; $n$
is 188 and 187 for the two Gemini swaps and 185 for its language-difference contrast. All-English refusal is
95.8\%, 97.4\%, 99.5\% and 88.4\% on the four models in table order.}
\label{tab:replication}
\footnotesize
\setlength{\tabcolsep}{4pt}
\begin{tabular}{@{}lrrrrr@{}}
\toprule
model & Spanish swap & $p_{\mathrm{Holm}}$ & Chinese swap & $p_{\mathrm{Holm}}$ & language difference \\
\midrule
gemini & \RefusalGemSwapSRd{} \RefusalGemSwapSCi & \RefusalGemSwapSPpre & \RefusalGemSwapZRd{} \RefusalGemSwapZCi & \RefusalGemSwapZPpre & \RefusalGemTransferRd{} \RefusalGemTransferCi \\
deepseek & \RefusalDsSwapSRd{} \RefusalDsSwapSCi & \RefusalDsSwapSPpre & \RefusalDsSwapZRd{} \RefusalDsSwapZCi & \RefusalDsSwapZPpre & \RefusalDsTransferRd{} \RefusalDsTransferCi \\
\midrule
gpt-5.4-mini & \RefusalgptminiGptminSwapSRd{} \RefusalgptminiGptminSwapSCi & \RefusalgptminiGptminSwapSPpre & \RefusalgptminiGptminSwapZRd{} \RefusalgptminiGptminSwapZCi & \RefusalgptminiGptminSwapZPpre & \RefusalgptminiGptminTransferRd{} \RefusalgptminiGptminTransferCi \\
mistral-s4 & \RefusalmistralsmallMistraSwapSRd{} \RefusalmistralsmallMistraSwapSCi & \RefusalmistralsmallMistraSwapSPpre & \RefusalmistralsmallMistraSwapZRd{} \RefusalmistralsmallMistraSwapZCi & \RefusalmistralsmallMistraSwapZPpre & \RefusalmistralsmallMistraTransferRd{} \RefusalmistralsmallMistraTransferCi \\
\bottomrule
\end{tabular}
\end{table*}

\texttt{gpt-5.4-mini} refuses 97.4\% to 99.5\% in every cell. Its swap estimates are
\RefusalgptminiGptminSwapSRd{} and \RefusalgptminiGptminSwapZRd\pp{}
(\RefusalgptminiGptminSwapSDisc{} and \RefusalgptminiGptminSwapZDisc{} discordant), neither
distinguishable from zero, and the one effect surviving its own correction is the Chinese
operation main effect at $-1.9\pp$, Gemini's direction at a quarter of the size. Its unsafe
label is at the floor, 5 positives in 2{,}691 scorable responses under the primary judge and 19
under the second with 1 shared ($\kappa = 0.081$ at 99.2\% raw agreement), so the nesting
reported for the primary analysis is not established here and its absolute unsafe rate should not be
ranked against the others. Mistral Small 4 refuses 73.7\% to 88.4\%. Its Spanish swap is
\RefusalmistralsmallMistraSwapSRd\pp{} on \RefusalmistralsmallMistraSwapSDisc{} discordant
pairs (exact $p < 10^{-4}$; \RefusalmistralsmallqwenMistraSwapSRd{} under the second judge),
its Chinese swap \RefusalmistralsmallMistraSwapZRd\pp{} (\RefusalmistralsmallMistraSwapZDisc),
and unsafe compliance mirrors both at \UnsafemistralsmallMistraSwapSRd{} and
\UnsafemistralsmallMistraSwapZRd\pp; the two judges agree ($\kappa = 0.954$ refusal, $0.917$
unsafe).

Across the four models, the Spanish swap estimates range from $0.5$ to $13.7\pp$ and the
Chinese estimates from $-0.5$ to $5.3\pp$. Only Mistral's Spanish swap survives the combined
twelve-test Holm correction. Its Chinese interval includes Gemini's $5.3\pp$ point estimate,
whereas the DeepSeek and GPT intervals do not; these marginal intervals do not constitute a
between-model test. Two of the four language-difference estimates are positive, and both exclude zero.
Mistral has the lowest all-English refusal and the largest Spanish estimate, a co-occurrence
across four models rather than an established relation between baseline refusal and swap size.
Its request-slot Chinese main effect is $-4.0\pp$ (Holm $0.060$ within that model's family).

\section{Discussion}
\label{sec:discussion}

\subsection{What component assignment adds}

A prompt's language inventory leaves component assignment unspecified. The paired reversal
measures a comparison that whole-prompt translation does not estimate. Model-language
patterns in Table~\ref{tab:replication} concern complete components, including their
text quantities. GPT-5.4-mini has refusal rates of 97.4--99.5\% across all nine configurations, providing a
high-refusal comparison to the larger Mistral contrast.

Evaluations should record component languages, match behaviours, target language and translated-span counts,
and report translated text quantity. Lower refusal warrants examining unsafe compliance
separately, since declining less often is not itself evidence of guardrail bypass. The
185-response audit's lower observed automatic-to-human agreement on unsafe labels reinforces this distinction
(Section~\ref{sec:human-audit}). Table~\ref{tab:protocol} records the measurement checks used here.

\begin{table}[!tb]
\centering
\caption{Measurement checks adopted in this study. Each check is one contingency table or one
paired difference, and each observation is from this corpus or its two-language pilot; they are
not a validated general protocol.}
\label{tab:protocol}
\scriptsize
\begin{tabular}{@{}p{0.44\columnwidth}p{0.49\columnwidth}@{}}
\toprule
check & observed sensitivity or exclusion \\
\midrule
Segment before translating; tabulate coverage by source-native group.
& A qualifier occurs on 39/53 copyright and 6/21 generic harmful behaviours in HarmBench. \\[2pt]
Count translated spans per configuration; match this count for component swaps.
& In different pilot strata, the unsafe swap was $+13.2\pp$ where span counts differed and
  $0.0\pp$ where they matched, using any-positive labels from the first two responses. \\[2pt]
Pin unmanipulated spans to the base language and count realised dose on the built dataset.
& Under the inheriting convention the contrast was dose-balanced on 52\% of behaviours; under
  pinning, on 305 of 305. \\[2pt]
Treat a generation with no text as missing, not as a non-refusal.
& Scoring 10 empty draws as \emph{refusal: no} inflated the headline contrast by $1.0$ and
  $1.5\pp$. \\[2pt]
Separate the naturalness floor (rating 1) from a fluency threshold; report exclusions and missing ratings by configuration.
& A naturalness floor of 3 keeps 95\% of the monolingual configuration and 44\% of a switched
  one. \\[2pt]
Use two judges from vendors supplying no evaluated model; check whether their positive sets
nest or cross.
& In the pilot a vendor swap moved the overall unsafe rate from 6.4\% to 17.4\%; here it moved
  the four unsafe-compliance component-swap contrasts by at most $1.1\pp$. \\
\bottomrule
\end{tabular}
\end{table}

\subsection{Explanations to test}

Uneven language exposure is one hypothesis for Figure~\ref{fig:grid}'s patterns.
Pretraining coverage could affect instruction or harmful-content recognition; language
coverage during safety fine-tuning could affect subsequent refusal. These are distinct,
untested mechanisms: training proportions were neither measured nor manipulated here.
Testing them requires language-resolved training information and controlled comparisons of
pretraining and safety-tuning coverage.

Translation quantity and quality are further candidate explanations. Operation translations
average 14.3 characters in Chinese against 5.3 for requests across the 305-behaviour build
(14.3 and 6.1 on the analysed 190). These \texttt{len(text)} counts include Latin letters,
digits, spaces and punctuation. Equal span counts leave unequal quantities; comparisons
matching translated quantity while changing the component would help separate these explanations.

\section{Limitations}
\label{sec:limitations}

\paragraph{Analysed population.} Results are for the 190 of
300 harmful behaviours with a two-of-three consensus on both boundaries (245) that then survive
the gate on all nine configurations. The automatic gate excludes whole behaviours with
negative or missing intent verdicts under the completeness rule, changing source-native
composition (Table~\ref{tab:flow}). Findings describe behaviours with separable components
whose composed versions pass this automatic gate. Copyright behaviours are absent from the
analysed set, and only three chemical/biological behaviours remain. The inherited sampling
strata do not support semantic harm-class coverage claims.

\paragraph{Three design choices bound the question.} With $s = 2$ slots, $k_{\max} = 2$, so
the ``more languages'' comparison has two points and is not a dose-response. The frame is
English in all nine configurations, so nothing here speaks to varying the matrix language. A
trailing qualifier stays English for the dose reason of Section~\ref{sec:dose}; this affects 28
of the 190 analysed behaviours (15\%), and on the other 85\% the operation slot carries the
whole harmful predicate.

\paragraph{Judge validity.} The 185-response audit supports agreement with a human reference
on the retained harmful population, with weaker agreement for unsafe compliance than refusal
(Section~\ref{sec:human-audit}). It excludes benign and reference responses and does not
validate the intent or naturalness construction gates. Two uncertain ratings and the few
original disagreements required author resolution; uncertainty intervals and weighting assumptions are given in Appendix~\ref{app:audit}. Automatic rates remain specific to this behaviour
set and rubric rather than comparable measures of realised harm across studies.

\paragraph{Four evaluated models, two non-English languages, one behaviour set.} Two of the
four models were added after the primary result was fixed and are corrected separately; the
extension adds model coverage on the same 190 behaviours, not an independent behaviour set.
Between-model patterns involving the two added models are descriptive; the direct comparison
between the two primary models is reported in Section~\ref{sec:results}. GPT-5.4-mini refuses more than 97\% of these behaviours in every cell. All composition used one translation
system, so a language effect is joint with how that system renders it. Chinese and Spanish are
both high-resource. Extending this design to lower-resource languages would require measuring
translation quality and intent preservation in those languages.

\section{Conclusion}

Component-language assignment adds a comparison to multilingual safety evaluation.
On the retained 190 behaviours, primary-analysis swaps fail Holm correction while the exploratory
Mistral Spanish contrast survives the twelve-test sensitivity correction. The varying
estimates support reporting component assignments alongside whole-prompt language, with
model- and language-specific comparisons. Evaluators should match behaviours and translated-span
counts, report translated text quantity, and interpret refusal separately from unsafe
compliance.

\section*{Acknowledgment}
OpenAI Codex assisted with prose revision throughout the manuscript and with preparing the
offline statistical replay and submission materials. It was not used to supply human
validation labels. The authors are responsible for the final text and its interpretation.

\bibliographystyle{IEEEtranN}
\bibliography{custom}

\appendices

\section{Source-native coverage and selection}
\label{app:coverage}

Table~\ref{tab:app-coverage} gives, for every source-native group in the 370-behaviour pool,
how many behaviours the rule-based chunker segments into both slots, how many carry an
explicit qualifier, and how many have a content word in the frame. Section~\ref{sec:design}
quotes the copyright, generic-harmful and cybercrime rows. Table~\ref{tab:app-crosswalk}
crosses the inherited sampling strata with the source-native groups at the three selection
stages of Table~\ref{tab:flow}; the inherited mapping that sends HarmBench \emph{copyright}
to \texttt{physical\_harm} and \emph{illegal} to \texttt{fraud\_deception} is visible in it.

\begin{table}[!tb]
\centering
\caption{Slot coverage by source-native group over the 370-behaviour pool. Two slots: the
rule-based chunker recovers both the request and the operation slot. Qualifier: an explicit
qualifier span is present. Frame content word: at least one content word lies outside the two
slots.}
\label{tab:app-coverage}
\footnotesize
\resizebox{\columnwidth}{!}{\AppCoverageTable}
\end{table}

\begin{table}[!tb]
\centering
\caption{Inherited sampling stratum against source-native group for the 300 sampled, 245
consensus and 190 analysed harmful behaviours.}
\label{tab:app-crosswalk}
\footnotesize
\resizebox{\columnwidth}{!}{\AppCrosswalkTable}
\end{table}

\section{Gate decisions by configuration}
\label{app:gate}

Table~\ref{tab:app-gate} counts, per configuration, the intent verdicts and naturalness
ratings behind the exclusions of Section~\ref{sec:design}. The two intent columns cover the
2{,}205 harmful factorial items of the \AppGateHarmfulSeeds{} consensus behaviours; the
paired Cochran's $Q$ statistics quoted there, \AppGateQMissing{} ($p = \AppGatePMissing$)
for missing and \AppGateQNegative{} ($p = \AppGatePNegative$) for negative verdicts, are
computed on these counts. The naturalness-1 column counts exclusions over all 3{,}355
composed items, benign included; the unrated column counts the 72 analysed harmful
factorial prompts without a naturalness rating, which are retained. Both naturalness
columns concentrate in the monolingual-slot configurations rather than in the four
single-switch configurations that carry the component-swap contrasts. All
\AppGateCanonicalNegatives{} negative verdicts on the canonical configuration \cfg{EE} fall
on copyright behaviours.

\begin{table}[!tb]
\centering
\caption{Intent verdicts and naturalness ratings by configuration. Intent columns: harmful
factorial items of the 245 consensus behaviours rated as losing intent (negative) or returned
without a verdict (missing). Naturalness 1: items excluded as not interpretable as a sentence,
all pools. Naturalness unrated: analysed harmful factorial prompts (190 behaviours) without a
rating.}
\label{tab:app-gate}
\footnotesize
\resizebox{\columnwidth}{!}{\AppGateTable}
\end{table}

\section{Round-trip diagnostic for isolated spans}
\label{app:roundtrip}

Table~\ref{tab:app-roundtrip} compares the round-trip similarity of a behaviour's whole-sentence
translation with the mean round-trip similarity of its spans translated in isolation, paired
on behaviour over the 305 composed behaviours. The difference is spans minus sentence, so a
negative value means isolated spans round-trip worse. The replay rows are computed from the
retained English source and back-translation strings with a BCa bootstrap
(\AppRoundtripResamples{} resamples, seed \AppRoundtripSeed); \AppRoundtripMissingEs{}
Spanish whole-sentence back-translations (eight benign items and one harmful) are no longer
available, so the Spanish replay uses 296 pairs. The original rows are the summary from the
run on all 305 pairs. Neither comparison detects a difference, and neither is an equivalence
test.

\begin{table}[!tb]
\centering
\caption{Whole-sentence against isolated-span round-trip similarity, paired on behaviour.
Sentence and spans are mean similarities; the interval is a 95\% bootstrap interval for the
paired difference; spans worse counts behaviours whose isolated spans score below their
sentence.}
\label{tab:app-roundtrip}
\scriptsize
\setlength{\tabcolsep}{3pt}
\resizebox{\columnwidth}{!}{\AppRoundtripTable}
\end{table}

\section{Human response-label audit}
\label{app:audit}

The audit sampled \AppAuditItems{} responses from the \AppAuditPopulation{} nonempty harmful
factorial responses of the four models, by simple random sampling without replacement within
the \AppAuditStrata{} model-by-configuration strata (five or six per stratum), covering
\AppAuditClusters{} behaviours. Two raters received the original English behaviour, the
complete response and the rubric of Section~\ref{sec:outcome}, in packets that hid the model,
the configuration and the automatic labels. Two unsafe-compliance ratings marked uncertain
are excluded from that label's denominator rather than counted as negative, and
\AppAuditAdjudicated{} unresolved labels on \AppAuditAdjudicatedItems{} responses were
settled by author adjudication, giving a complete reference with \AppAuditRefRefusalYes{}
refusals and \AppAuditRefUnsafeYes{} unsafe responses among the \AppAuditItems.

Table~\ref{tab:app-audit} gives the agreement tables between the two raters' original labels
and between the resolved reference and each automatic judge. In each row the first-named
party is read first: YN counts responses the first party labelled positive and the second
negative. Of the judges' unsafe-compliance disagreements with the reference, six of seven
for the primary judge and seven of eight for the second are responses the judge marks
unsafe and the reference does not; the remaining one in each case is the reverse. Cluster
intervals are percentile intervals from a bootstrap over the \AppAuditClusters{} observed
behaviour clusters, moving all sampled responses of a behaviour together, with
\AppAuditResamples{} resamples and seed \AppAuditSeed; the weighted column reweights each
response by its stratum's inclusion probability ($N_h/n_h$). The intervals approximate the
sampling design and are not a full stratified finite-population variance.

\begin{table}[!tb]
\centering
\caption{Agreement on the \AppAuditItems-response audit. Counts are two-by-two tables
(YY both positive, YN first positive and second negative, NY the reverse, NN both negative);
the unsafe-compliance rater comparison has 183 pairs after excluding two uncertain ratings.
$\kappa$ is Cohen's kappa with an approximate behaviour-cluster 95\% percentile interval;
weighted $\kappa$ uses the design weights.}
\label{tab:app-audit}
\scriptsize
\setlength{\tabcolsep}{3pt}
\resizebox{\columnwidth}{!}{\AppAuditTable}
\end{table}

\section{Per-configuration rates and further swap contrasts}
\label{app:cells}

Figure~\ref{fig:grid} shows primary-judge refusal rates per configuration.
Table~\ref{tab:app-unsafe-cells} gives the corresponding unsafe-compliance rates under the
primary judge, Table~\ref{tab:app-refusal-qwen-cells} the refusal rates under the second
judge, each with Wilson 95\% intervals, and Table~\ref{tab:app-swaps} the component-swap and
language-difference contrasts under the second judge for refusal and under both judges for
unsafe compliance, in the layout of Table~\ref{tab:replication}. Rows in the cell tables are
grouped by dose, the number of translated spans: \cfg{EE} at dose 0, the four single-switch
configurations at dose 1 and the four double-switch configurations at dose 2.

\begin{table}[!tb]
\centering
\caption{Unsafe-compliance rates (\%) per configuration under the primary judge, with Wilson
95\% intervals; 190 harmful behaviours per cell except the Gemini cells with empty generations.}
\label{tab:app-unsafe-cells}
\scriptsize
\setlength{\tabcolsep}{2.5pt}
\resizebox{\columnwidth}{!}{\AppUnsafeCells}
\end{table}

\begin{table}[!tb]
\centering
\caption{Refusal rates (\%) per configuration under the second judge (\texttt{qwen3.8-flash}),
with Wilson 95\% intervals.}
\label{tab:app-refusal-qwen-cells}
\scriptsize
\setlength{\tabcolsep}{2.5pt}
\resizebox{\columnwidth}{!}{\AppRefusalSecondCells}
\end{table}

\begin{table}[!tb]
\centering
\caption{Component-swap contrasts (request-switched minus operation-switched at dose 1, in
percentage points with 95\% bootstrap intervals) and the language-difference contrast
(Spanish swap minus Chinese swap) under the second judge for refusal and under both judges
for unsafe compliance. $p_{\mathrm{Holm}}$ uses the same families as
Table~\ref{tab:replication}: the six pre-specified tests for the two primary models and a
three-test family for each extension model, each outcome and judge corrected separately.}
\label{tab:app-swaps}
\scriptsize
\setlength{\tabcolsep}{2.5pt}
\resizebox{\columnwidth}{!}{%
\begin{tabular}{@{}lrrr@{}}
\toprule
model & Spanish swap & Chinese swap & language difference \\
\midrule
\multicolumn{4}{@{}l}{\emph{refusal, second judge}} \\
Gemini 2.5 Flash & \RefusalqwenGemSwapSRd{} \RefusalqwenGemSwapSCi & \RefusalqwenGemSwapZRd{} \RefusalqwenGemSwapZCi & \RefusalqwenGemTransferRd{} \RefusalqwenGemTransferCi \\
 & $p_{\mathrm{Holm}}$ \RefusalqwenGemSwapSPpre & $p_{\mathrm{Holm}}$ \RefusalqwenGemSwapZPpre & \\
\addlinespace[1pt]
DeepSeek & \RefusalqwenDsSwapSRd{} \RefusalqwenDsSwapSCi & \RefusalqwenDsSwapZRd{} \RefusalqwenDsSwapZCi & \RefusalqwenDsTransferRd{} \RefusalqwenDsTransferCi \\
 & $p_{\mathrm{Holm}}$ \RefusalqwenDsSwapSPpre & $p_{\mathrm{Holm}}$ \RefusalqwenDsSwapZPpre & \\
\addlinespace[1pt]
GPT-5.4-mini & \RefusalgptminiqwenGptminSwapSRd{} \RefusalgptminiqwenGptminSwapSCi & \RefusalgptminiqwenGptminSwapZRd{} \RefusalgptminiqwenGptminSwapZCi & \RefusalgptminiqwenGptminTransferRd{} \RefusalgptminiqwenGptminTransferCi \\
 & $p_{\mathrm{Holm}}$ \RefusalgptminiqwenGptminSwapSPpre & $p_{\mathrm{Holm}}$ \RefusalgptminiqwenGptminSwapZPpre & \\
\addlinespace[1pt]
Mistral Small 4 & \RefusalmistralsmallqwenMistraSwapSRd{} \RefusalmistralsmallqwenMistraSwapSCi & \RefusalmistralsmallqwenMistraSwapZRd{} \RefusalmistralsmallqwenMistraSwapZCi & \RefusalmistralsmallqwenMistraTransferRd{} \RefusalmistralsmallqwenMistraTransferCi \\
 & $p_{\mathrm{Holm}}$ \RefusalmistralsmallqwenMistraSwapSPpre & $p_{\mathrm{Holm}}$ \RefusalmistralsmallqwenMistraSwapZPpre & \\
\midrule
\multicolumn{4}{@{}l}{\emph{unsafe compliance, primary judge}} \\
Gemini 2.5 Flash & \UnsafeGemSwapSRd{} \UnsafeGemSwapSCi & \UnsafeGemSwapZRd{} \UnsafeGemSwapZCi & \UnsafeGemTransferRd{} \UnsafeGemTransferCi \\
 & $p_{\mathrm{Holm}}$ \UnsafeGemSwapSPpre & $p_{\mathrm{Holm}}$ \UnsafeGemSwapZPpre & \\
\addlinespace[1pt]
DeepSeek & \UnsafeDsSwapSRd{} \UnsafeDsSwapSCi & \UnsafeDsSwapZRd{} \UnsafeDsSwapZCi & \UnsafeDsTransferRd{} \UnsafeDsTransferCi \\
 & $p_{\mathrm{Holm}}$ \UnsafeDsSwapSPpre & $p_{\mathrm{Holm}}$ \UnsafeDsSwapZPpre & \\
\addlinespace[1pt]
GPT-5.4-mini & \UnsafegptminiGptminSwapSRd{} \UnsafegptminiGptminSwapSCi & \UnsafegptminiGptminSwapZRd{} \UnsafegptminiGptminSwapZCi & \UnsafegptminiGptminTransferRd{} \UnsafegptminiGptminTransferCi \\
 & $p_{\mathrm{Holm}}$ \UnsafegptminiGptminSwapSPpre & $p_{\mathrm{Holm}}$ \UnsafegptminiGptminSwapZPpre & \\
\addlinespace[1pt]
Mistral Small 4 & \UnsafemistralsmallMistraSwapSRd{} \UnsafemistralsmallMistraSwapSCi & \UnsafemistralsmallMistraSwapZRd{} \UnsafemistralsmallMistraSwapZCi & \UnsafemistralsmallMistraTransferRd{} \UnsafemistralsmallMistraTransferCi \\
 & $p_{\mathrm{Holm}}$ \UnsafemistralsmallMistraSwapSPpre & $p_{\mathrm{Holm}}$ \UnsafemistralsmallMistraSwapZPpre & \\
\midrule
\multicolumn{4}{@{}l}{\emph{unsafe compliance, second judge}} \\
Gemini 2.5 Flash & \UnsafeqwenGemSwapSRd{} \UnsafeqwenGemSwapSCi & \UnsafeqwenGemSwapZRd{} \UnsafeqwenGemSwapZCi & \UnsafeqwenGemTransferRd{} \UnsafeqwenGemTransferCi \\
 & $p_{\mathrm{Holm}}$ \UnsafeqwenGemSwapSPpre & $p_{\mathrm{Holm}}$ \UnsafeqwenGemSwapZPpre & \\
\addlinespace[1pt]
DeepSeek & \UnsafeqwenDsSwapSRd{} \UnsafeqwenDsSwapSCi & \UnsafeqwenDsSwapZRd{} \UnsafeqwenDsSwapZCi & \UnsafeqwenDsTransferRd{} \UnsafeqwenDsTransferCi \\
 & $p_{\mathrm{Holm}}$ \UnsafeqwenDsSwapSPpre & $p_{\mathrm{Holm}}$ \UnsafeqwenDsSwapZPpre & \\
\addlinespace[1pt]
GPT-5.4-mini & \UnsafegptminiqwenGptminSwapSRd{} \UnsafegptminiqwenGptminSwapSCi & \UnsafegptminiqwenGptminSwapZRd{} \UnsafegptminiqwenGptminSwapZCi & \UnsafegptminiqwenGptminTransferRd{} \UnsafegptminiqwenGptminTransferCi \\
 & $p_{\mathrm{Holm}}$ \UnsafegptminiqwenGptminSwapSPpre & $p_{\mathrm{Holm}}$ \UnsafegptminiqwenGptminSwapZPpre & \\
\addlinespace[1pt]
Mistral Small 4 & \UnsafemistralsmallqwenMistraSwapSRd{} \UnsafemistralsmallqwenMistraSwapSCi & \UnsafemistralsmallqwenMistraSwapZRd{} \UnsafemistralsmallqwenMistraSwapZCi & \UnsafemistralsmallqwenMistraTransferRd{} \UnsafemistralsmallqwenMistraTransferCi \\
 & $p_{\mathrm{Holm}}$ \UnsafemistralsmallqwenMistraSwapSPpre & $p_{\mathrm{Holm}}$ \UnsafemistralsmallqwenMistraSwapZPpre & \\
\bottomrule
\end{tabular}}
\end{table}

\end{document}